\documentclass{article}

\usepackage[preprint]{neurips_2026}

\usepackage[utf8]{inputenc} 
\usepackage[T1]{fontenc}    
\usepackage{hyperref}       
\usepackage{url}            
\usepackage{booktabs}       
\usepackage{amsfonts}       
\usepackage{nicefrac}       
\usepackage{microtype}      
\usepackage{xcolor}         
\usepackage{float}
\usepackage{comment}
\usepackage{pgfplots}
\pgfplotsset{compat=1.18}
\usepackage{wrapfig}
\usepackage{booktabs}
\usepackage{tcolorbox}
\usepackage{enumitem}
\usepackage{amsmath}
\usepackage{subcaption}
\usepackage[table]{xcolor}
\usepackage[hybrid]{markdown}
\definecolor{encodingblue}{HTML}{0072B2}
\definecolor{decodingorange}{HTML}{E69F00}

\title{Repurposing Pre-trained LLMs as High Fidelity Continuous Text Autoencoders}

\author{%
  Arkanath Pathak \hspace{1.5cm} Unnat Jain \hspace{1.5cm} Alexander C. Berg \\
  University of California, Irvine \\
  \texttt{\{arkanatp, unnatj, bergac\}@uci.edu}
}

\begin{document}

\maketitle

\begin{abstract}
Next-token prediction has enabled highly fluent autoregressive language models, but it represents global structure only indirectly through sequential factorization. In contrast, high-fidelity autoencoders have become a standard primitive in image generation, enabling generative models to operate over continuous latent spaces; text lacks a comparably faithful continuous representation. We propose LLMAE, a method for repurposing a pretrained decoder-only language model as a continuous text autoencoder by exposing the activations of an intermediate layer as a fixed-length latent bottleneck. LLMAE achieves this interface with structured attention masks and LoRA adaptation, leveraging the generative prior of the original LLM. Across two backbones (270M Gemma 3 and 0.5B Qwen2.5), LLMAE reconstructs sequences up to 1024 tokens with high accuracy, reproducing up to 97\% of documents verbatim. We demonstrate downstream utility by training a lightweight diffusion model that generates detailed image captions directly in the frozen LLMAE latent space. By mapping text into a fixed-length continuous latent space, our approach provides an effective substrate for downstream adaptation while benefiting from the fluency of the original LLM. Code and models are available at \href{https://github.com/arkanath/LLMAE}{https://github.com/arkanath/LLMAE}.
\end{abstract}

\section{Introduction}

Autoregressive LLMs dominate text generation, demonstrating remarkable efficacy and fluency, in part by capitalizing on the inherently sequential nature of language. However, this strict token-by-token factorization necessitates local decoding decisions, fundamentally constraining the model's capacity for global foresight and planning \citep{yao2023tree}. Furthermore, downstream adaptation remains expensive and non-trivial \cite{hu2022lora}, as traditional fine-tuning often induces ``catastrophic forgetting'' \cite{kotha2023understanding} of the LLM's original generative prior \cite{holtzman2019curious}. Non-autoregressive (NAR) generation has been studied to circumvent this sequential bottleneck, yet these methods often struggle to model word-order dependencies explicitly, leading to text that lacks the grammatical correctness and fluency characteristic of autoregressive models \cite{li2023diffusion}. A prominent framework for NAR generation is text diffusion, however, token-level diffusion remains non-trivial, requiring specific pre-training objectives and carefully designed training/sampling procedures \cite{lou2023discrete, dieleman2022continuous}.
 
\begin{figure*}[t]
    \centering
    \begin{subfigure}{0.36\textwidth}
        \centering
        \includegraphics[width=\linewidth, trim={15pt 0 0 0}, clip]{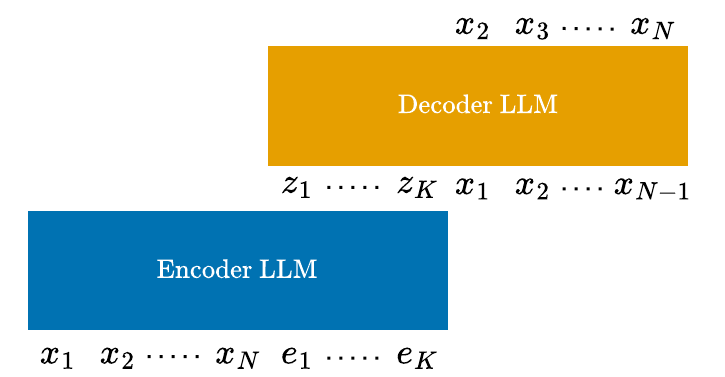}
        \caption{Soft Prompting}
        \label{fig:arch_softprompt}
    \end{subfigure}
    \hfill
    \begin{subfigure}{0.26\textwidth}
        \centering
        \includegraphics[width=0.75\textwidth]{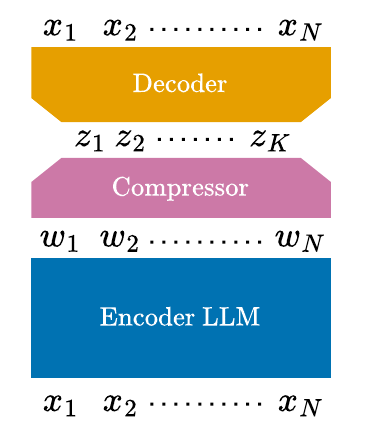}
        \caption{Token Compressor}
        \label{fig:arch_project}
    \end{subfigure}
    \hfill
    \begin{subfigure}{0.36\textwidth}
        \centering
        \includegraphics[width=\linewidth, trim={0 8pt 20pt 0}]{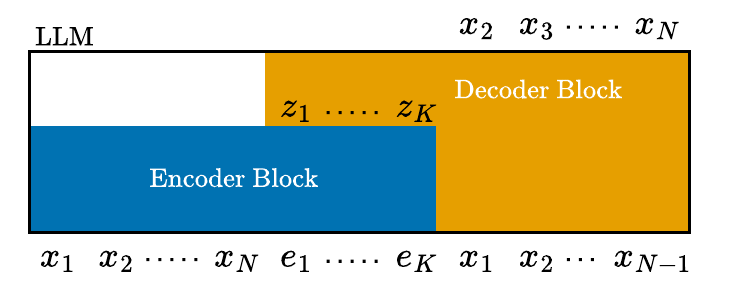}
        \caption{LLMAE (Ours)}
        \label{fig:arch_llmae}
    \end{subfigure}

    \caption{\textbf{Related LLM-based autoencoding architectures.} (a)~Soft prompting: an LLM encodes input tokens $x_1 \ldots x_N$ with learnable query tokens $e_1 \ldots e_K$; its activations at $e$ form the latent $z_1 \ldots z_K$, an input for a separate LLM. (b)~Token compressor: output token embeddings are compressed into $z_1 \ldots z_K$ by a separate module. (c)~LLMAE (ours): LLM split by attention masks into an Encoder Block and a Decoder Block; $z_1 \ldots z_K$ are the activations of $e$ at an intermediate layer.}

    \label{fig:architecture_comparison}
\end{figure*}

Drawing inspiration from the visual domain, where high-fidelity autoencoders are a standard primitive \cite{rombach2022high}, we propose LLMAE (LLM AutoEncoder), a faithful text autoencoder that maps sequences of any length into a fixed-length continuous representation. Non-autoregressive generative models are built to operate on fixed-shape tensors, and freezing the autoencoder enables easier downstream modeling without the cost or catastrophic forgetting of token-level fine-tuning. LLMAE partitions a single lightweight LLM into an integrated encoder and decoder (Fig.~\ref{fig:arch_llmae}) by utilizing structured attention masks and utilizing LoRA \cite{hu2022lora} for lightweight fine-tuning. Specifically, activations at an intermediate transformer layer aggregate abstract information to serve as latent encodings, while subsequent layers function as a fluent decoder. This enables high-fidelity reconstruction while benefiting from the fluency of the original model. We also propose an optional single-layer, identity-initialized additive codec over the latent sequence, which learns a locally smooth latent interface under KL regularization while improving the reconstruction quality further. We instantiate LLMAE on two lightweight backbones from different model families, Gemma-3-270M \cite{gemmateam2025gemma3technicalreport} and Qwen2.5-0.5B \cite{qwen2p5}, training only 10.7M of 278.8M and 23.8M of 494M parameters. Our main contributions in this paper are the following:

\textbf{LLMAE Framework:} We introduce an approach for repurposing pre-trained LLMs as text autoencoders by establishing a fixed-length continuous latent bottleneck at an intermediate layer.

\textbf{Reconstruction Analysis:} We demonstrate near-perfect word-level fidelity on text sequences up to 1024 tokens long ($\approx$ 4.6K characters or 760 words), with up to $97.4\%$ of documents reproduced verbatim under strict exact match, significantly outperforming previous text autoencoders. We provide detailed ablations of latent depth, latent capacity, and the KL/codec components for two backbones, identifying the intermediate-layer bottleneck as the core enabling factor and confirming the ``entropy valley'' \cite{skean2025layer} as its best depth.

\textbf{Downstream Generative Utility:} We train a latent text diffusion model for detailed image captioning on frozen LLMAE latents. Our lightweight diffusion model (111M trainable parameters) generates fluent, high-quality captions. We remain behind frontier VLMs, which fine-tune multi-billion-parameter backbones on far larger corpora, but outperform some substantially larger autoregressive captioners such as LLaVA-1.5-7B \cite{liu2024improved} and Share-Captioner \cite{chen2024sharegpt4v}.

\section{Related Work}
We discuss lines of work related to this paper: context compression in LLMs, text autoencoding architectures, evidence for what different LLM layers encode, and diffusion modeling for text.

{\bf Context Compression.} We establish a bottleneck in an intermediate layer of the LLM and use that bottleneck as the latent representation. This is similar to work on context compression, which adapts pre-trained LLMs to condense long prompts into compact representations. Early methods demonstrated that learnable continuous tokens, whether only at the input layer via Soft Prompting \cite{lester2021power} or also integrated into intermediate layers via Prefix Tuning \cite{li2021prefix}, can compress context and steer model behavior. This concept evolved into structural techniques that cache LLM internal activations for compression, {\em e.g.}, Gist Tokens \cite{mu2023learning}. The standard causal attention mask is modified so that query tokens can only attend as far back as the activations for the Gist tokens, forming a bottleneck that effectively represents context but has trouble scaling to longer sequences without architecture modification \cite{petrov2025long}. ICAE \cite{ge2023context} fine-tunes an LLM to compress context into soft prompts that are decoded by a frozen LLM for reconstruction and task completion.  Our experiments show that using activations of an intermediate layer provides better reconstruction quality and is achievable using a single lighter weight LLM.

{\bf Text Autoencoder Architectures.} Approaches in related recent literature on text autoencoder architecture generally fall into two categories. The ``soft prompt'' approaches (Fig.~\ref{fig:arch_softprompt}) can use a variety of models for encoder and decoder.  ICAE \cite{ge2023context} fine-tunes an LLM with LoRA as an encoder and uses a fixed LLM as the decoder.  PLANNER \cite{zhang2023planner} takes the first K activations from the final layer of a fine-tuned BERT encoder as the latent and uses an autoregressive GPT-2 decoder to decode back to text.

The other category of approach (Fig.~\ref{fig:arch_project}) compresses LLM output activations into latents for diffusion. LD4LG \cite{lovelace2023latent} repurposes pretrained encoder-decoder backbones (such as BART, \citealp{lewis2020bart}) by utilizing a Perceiver \cite{jaegle2021perceiver} style query-based compression mechanism to gather information into a fixed-length latent sequence. These compressed tokens are then used to condition the autoregressive decoder for short-form text generation. COSMOS \cite{meshchaninov2025cosmos} follows a related approach, taking semantically rich hidden representations from a frozen BERT encoder and utilizing a Perceiver Resampler architecture \cite{alayrac2022flamingo} to compress the variable-length sequence into a fixed-size latent matrix. This fixed-size matrix is later expanded by a symmetric decompressor back into corresponding BERT tokens, which are then linearly projected into vocabulary probabilities for token prediction.

Our approach (Fig.~\ref{fig:arch_llmae}) partitions a single LLM for both encoding and decoding, simplifying modeling compared to soft-prompting, and uses the same LLM to reduce the sequence length instead of adding on a separate compressor and decoder.  

{\bf LLM Representations.}  Research exploring the characteristics of LLM layers suggests the existence of an ``entropy valley,'' where autoregressive models exhibit a distinct dip in entropy and a peak in representational accuracy at intermediate layers \cite{skean2025layer,lad2024remarkable}. LLMAE builds on this finding by placing its bottleneck at an intermediate layer, and our depth ablation (Table~\ref{tab:depth_ablation}) supports it: deep intermediate layers carry the information a bottleneck needs in an accessible form.

{\bf Text Diffusion.} Diffusion for text is in part inspired by the success of Diffusion models \cite{ho2020denoising} for images where it often operates on continuous latent representations obtained from pre-trained autoencoders \cite{rombach2022high}, where recent work shows potential advantages for more semantically grounded encoders \cite{zheng2025diffusion}. Text diffusion can work via discrete noising processes on tokens, \cite{shi2024simplified, nie2025large, von2025generalized, lou2023discrete}, by Gaussian noise on continuous token embeddings \cite{li2022diffusion, dieleman2022continuous, shabalin2025tencdm}, or by Gaussian noise on latent embeddings that encode multi-token or long-form text \cite{lovelace2023latent, zhang2023planner, lovelace2024diffusion, meshchaninov2025cosmos}. Diffused latent embeddings are commonly decoded autoregressively for fluent output, but most such work has only targeted short texts. Challenges include achieving low latent-to-text conversion error and smoothness in the latent distribution as highlighted in \cite{zhang2023planner, meshchaninov2025cosmos}. Learning latent representations that can effectively encode long passages and work for diffusion is still an open challenge \cite{arriola2025block} that we address in this paper.

\section{Method}
\begin{figure}[t]
    \centering
    \includegraphics[width=\textwidth, trim={20pt 0 0 0}]{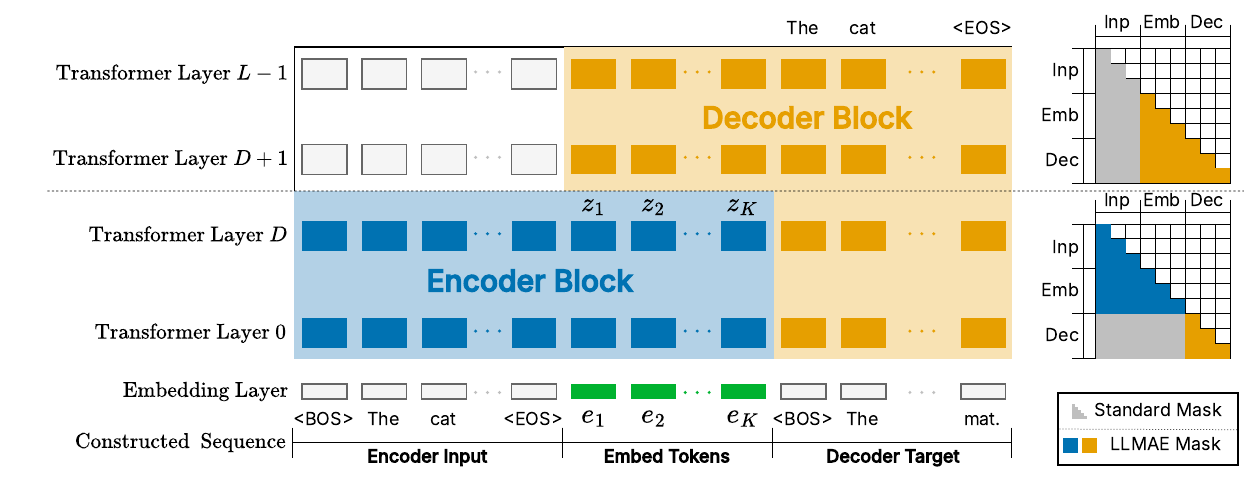}
    \vspace{-15pt}
    \caption{\textbf{Latent bottleneck via attention masking.} Tokens of one color form a causal segment attending only to the matching segment of the previous layer. Encoder block (\textcolor{encodingblue}{blue}): Embed tokens attend to Input. Decoder block (\textcolor{decodingorange}{orange}): Output tokens attend only to the latents $z_i$ and previously decoded Output tokens. This establishes an autoencoding flow within an autoregressive backbone.}
    \label{fig:full_attention_masking}
\end{figure}

LLMAE turns a pretrained LLM into a text autoencoder. Its input is a text of up to $N$ tokens; its output is a fixed-length latent $\mathbf{z}$ of $K$ vectors, which the same LLM decodes back into the text. To obtain $\mathbf{z}$, we append $K$ learnable ``Embed'' tokens to the input and read their activations at an intermediate layer $D$. A modified attention mask splits the network there: the layers up to $D$ serve as the encoder, and the layers after $D$ serve as the decoder, which can attend to $\mathbf{z}$ but to nothing else about the input, so the reconstruction must pass through the latent (Fig.~\ref{fig:full_attention_masking}). The LLM is adapted with LoRA, and a small additive codec refines $\mathbf{z}$. The rest of this section describes the model, the codec, the training objective, and the training details.

\paragraph{Model Details.} We partition the depth $L$ of the model into two functional blocks: an Encoder Block ($0 \le l \le D$) and a Decoder Block ($D < l \le L-1$). At the transition layer $D$, the activations corresponding to the $K$ sequence positions where the learnable Embed tokens $e$ were placed define the latent representation $\mathbf{z} \in \mathbb{R}^{K \times d}$. Therefore, $\mathbf{z}$ represents the transformation of the original Embed tokens through the depth of the Encoder Block. At layer $D+1$ within the subsequent Decoder Block, the masking constraints are relaxed to allow the Output tokens to attend to the resulting latent states $\mathbf{z}$ while maintaining a strict mask against the original Input sequence. Since the Output tokens always remain isolated from the Input tokens, this configuration ensures that the reconstruction process is channeled through the latent bottleneck. For more details, we specify the attention mask modification in Table~\ref{tab:masking} in App.~\ref{sec:sup_attention_masking_strategy}. Formally, this partition can be expressed as:
$$\mathbf{z} = \text{Encoder}(\mathbf{x}); \text{  } \hat{\mathbf{x}} = \text{Decoder}(\mathbf{z}, \hat{\mathbf{x}}_{<i}),$$
where $\hat{\mathbf{x}}_{<i}$ denotes the tokens already decoded. For encoding and downstream modeling, $\mathbf{z}$ can be cached as the latent representation of the input. At inference time for decoding, we inject the latent $\mathbf{z}$ as input to layer $D+1$ corresponding to the Embed tokens, resulting in an inference-time complexity comparable to standard LLM decoding.

\paragraph{Additive Codec Refinement.}
LoRA adapts the backbone through a low-rank update to its attention projections in every layer. This keeps adaptation cheap and steers the generative prior, but it proves challenging to transform the latent's coordinate statistics. That matters on Gemma, whose raw latent inherits the backbone's activation geometry, with a few coordinates orders of magnitude larger than the rest (Table~\ref{tab:latent_smoothness}). We therefore introduce a small codec that acts on the latent alone. It consists of single-layer Transformer blocks ($\mathcal{T}_{\text{enc}}, \mathcal{T}_{\text{dec}}$) with per-dimension learnable scalar multipliers ($\boldsymbol{\alpha} \in \mathbb{R}^{d}$) initialized to zero, so that it starts as the identity and learns only a residual correction:
$$\mathbf{z}_{\text{codec}} = \mathbf{z} + \boldsymbol{\alpha}_{\text{enc}} \odot \mathcal{T}_{\text{enc}}(\mathbf{z}); \quad \hat{\mathbf{z}} = \mathbf{z}_{\text{codec}} + \boldsymbol{\alpha}_{\text{dec}} \odot \mathcal{T}_{\text{dec}}(\mathbf{z}_{\text{codec}})$$
The multipliers are shared across all $K$ tokens but learned per hidden dimension $d$. We enforce the replacement $\mathbf{z} \leftarrow \hat{\mathbf{z}}$ to ensure the decoder conditions on the refined latent, stashing the detached $\mathbf{z}_{\text{codec}}$ as the definitive latent representation for KL regularization during training and downstream inference. Because the codec is additive by design, we finetune it jointly over the LLMAE backbone.

\subsection{Training Objectives}
The reconstruction task is formulated by constructing a composite sequence $\mathbf{s}$ that concatenates the input source, the introduced Embed tokens, and the reconstruction target. Let $\mathbf{x} = \{x_1, \dots, x_N\}$ denote the input sequence and $\mathbf{e} = \{e_1, \dots, e_K\}$ represent the learnable Embed tokens. Shorter sequences are padded to reach the $N$ token limit. The full training sequence is constructed as:$$\mathbf{s} = [\mathbf{x}_{\text{enc}}] \parallel [\mathbf{e}] \parallel [\mathbf{x}_{\text{dec}}]$$where $\mathbf{x}_{\text{enc}}$ and $\mathbf{x}_{\text{dec}}$ are identical sequences of tokens $x_i \in \mathcal{V}$. The total loss is a weighted sum:
$$\mathcal{L}_{\text{LLMAE}} = \mathcal{L}_{\text{NTP}} + \lambda_{z}\,\mathcal{L}_{z} + \lambda_{\text{ref}}\,\mathcal{L}_{\text{ref}},$$

where $\mathcal{L}_{\text{NTP}}$ is the next-token prediction cross-entropy loss of the decoder over the $N$ reconstruction positions $\mathbf{x}_{\text{dec}}$; $\mathcal{L}_{z} = D_{\text{KL}}\big(q(\mathbf{z}) \,\|\, \mathcal{N}(\mathbf{0},\mathbf{I})\big)$ is the KL divergence of the latent from a standard normal prior, with $q$ a Gaussian fitted per latent coordinate across the batch (Eq.~\ref{eq:kl_divergence} in App.~\ref{sec:supp_latent_smoothness}); $\mathcal{L}_{\text{ref}} = D_{\text{KL}}\big(\pi_{\theta} \,\|\, \pi_{\text{ref}}\big)$ is the KL divergence of the adapted model's next-token distribution $\pi_{\theta}$ from that of the frozen pretrained backbone LLM $\pi_{\text{ref}}$, before any adaptation, summed over the full vocabulary at each reconstruction position; and $\lambda_{z}$, $\lambda_{\text{ref}}$ are hyperparameters. In the codec stage, $\mathcal{L}_{z}$ is applied to $\mathbf{z}_{\text{codec}}$. Gradients of $\mathcal{L}_{\text{NTP}}$ flow back through the decoding layers into the latent and through the entire encoding block and the Embed tokens. The latent prior term regularizes the latent representations into a smooth manifold. The reference term encourages linguistic fluency by anchoring the LLMAE decoder to the pretrained backbone; $\pi_{\text{ref}}$ is evaluated with a forward pass on the target output tokens alone, without the input or the Embed tokens.

\begin{figure}[t]
    \centering
    \begin{subfigure}[b]{0.46\textwidth}
        \centering
        \includegraphics[width=\textwidth, trim={0pt 8pt 20pt 0}]{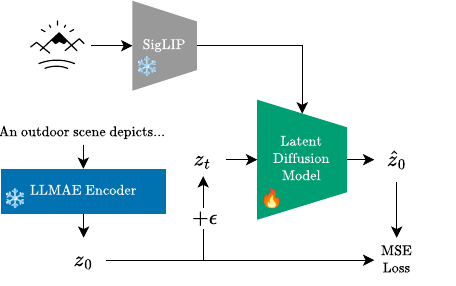}
        \caption{Training}
        \label{fig:training}
    \end{subfigure}
    \hfill
    \begin{subfigure}[b]{0.52\textwidth}
        \centering
        \includegraphics[width=\textwidth, trim={10pt 8pt 0 0}]{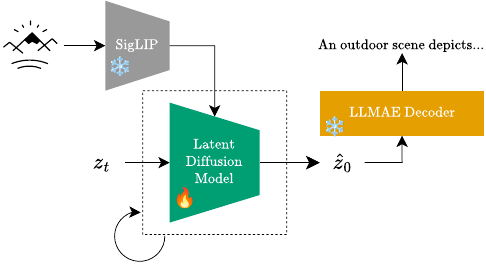}
        \caption{Inference}
        \label{fig:inference}
    \end{subfigure}
    \vspace{-5pt}
    \caption{\textbf{Latent Diffusion for Captioning.} A lightweight denoising score network conditioned on frozen SigLIP image features through cross-attention. LLMAE remains frozen throughout.}
    \label{fig:captioning_architecture}
\end{figure}

\subsection{Training Details}
We employ $N{=}1024$ input tokens and $K{=}256$ Embed tokens on both backbones. On Gemma we found it helpful to adopt a curriculum schedule, sorting samples by ascending length to stabilize early convergence before longer reconstruction; Qwen instead trains best on shuffled data (App.~\ref{sec:supp_curriculum_ablation}). We trained on 8 NVIDIA RTX PRO 6000 Blackwell GPUs; the full training takes $\approx 22$ hours. Full hyperparameters are listed in App.~\ref{sec:supp_architecture}.

\textbf{Training Data.} We train on a curated corpus of 1.4M samples drawn from two sources: 1M from Pile-uncopyrighted \cite{biderman2023pythia} for general web text and 400K from C4-RealNewsLike \cite{raffel2020exploring} for higher quality prose. We collect each source stratified by length for a balanced training set: documents are binned into 41 buckets of 100 characters.

\section{Experiments}
Our experiments focus on three core objectives: reconstruction fidelity, representational smoothness, and downstream generative utility. We first evaluate LLMAE’s autoencoding performance against existing baselines (Sec.~\ref{sec:eval_reconstruction}). We then analyze the distribution of the latent space (Sec.~\ref{sec:eval_smoothness}), evaluating the impact of KL regularization and the additive codec, before conducting systematic ablations on the latent bottleneck (Sec.~\ref{sec:eval_bottleneck_ablation}). Finally, we demonstrate the representation's utility (Sec.~\ref{sec:eval_captioning}) by training a latent diffusion model for detailed image captioning.

\subsection{Reconstruction Quality}
\subsubsection{Baselines}
We compare against two publicly released text autoencoders: ICAE \cite{ge2023context}, using its \texttt{v2} Stage~1 checkpoint for autoencoding, and COSMOS \cite{meshchaninov2025cosmos}, whose released checkpoint encodes up to 512 tokens into 512 latent tokens (no sequence-length reduction) and was not trained on longer inputs.
\subsubsection{Evaluation Datasets}
We evaluate on three held-out test sets of samples drawn from different text domains to assess generalization across writing styles and content types. All three sources are disjoint from the training corpus: C4-News-Stratified is deduplicated by content hash against all 1.4M training samples, OpenWebText uses the held-out test split shared by COSMOS, and the CreationMMBench references are GPT-4o-generated text from an external benchmark. The full construction protocol is detailed in App.~\ref{sec:supp_architecture}. Since the COSMOS baseline \cite{meshchaninov2025cosmos} is limited to a 512-token length, we report results on two test sets per source in Table~\ref{tab:reconstruction}: a medium-length set (up to 512 tokens) and a long-length set (up to 1024 tokens). For calibration, 1024 tokens corresponds to roughly 4{,}600 characters or 760 words of English prose. Each setting is evaluated on 500 held-out samples, with the exception of CreationMMBench at 512 tokens (311 samples, the subset of the benchmark that fits the limit).

\paragraph{C4-News-Stratified.}
We sample from the C4 \cite{raffel2020exploring} corpus with the \textit{RealNewsLike} filter for news-like domains, stratified into the same 41 character-length buckets used for training so that the evaluation spans short through full-length inputs evenly; we refer to this set as C4-News-Stratified. Besides Table~\ref{tab:reconstruction}, this is the test set used to measure reconstruction quality in this paper.
\paragraph{OpenWebText.}
OpenWebText \cite{Gokaslan2019OpenWeb} is an open-source reproduction of the WebText dataset used to train GPT-2. We sample from the specific held-out test split shared by COSMOS \cite{meshchaninov2025cosmos}, which allows us to evaluate our model on the distribution used to train the COSMOS autoencoder checkpoint.
\paragraph{CreationMMBench.}
CreationMMBench \cite{fang2025creation} offers a test benchmark of image-grounded long-form text generation tasks spanning four categories: Literary Writing, Common Functional Writing, Professional Functional Writing, and Creative Multimodal Understanding. We evaluate on the \texttt{reference\_answer\_by\_gpt4o} field from the benchmark dataset, generated by GPT-4o. 

\subsubsection{Metrics}
LLMAE is intended as a near-lossless interface: downstream models consume the latent in place of the text. We therefore report exact match (EM) and word edit distance (W-ED) as direct fidelity measures, alongside BLEU-4 for comparability with prior autoencoding work. BLEU-4 above $0.99$ leaves little headroom and does not separate a perfect reconstruction from one with a single wrong word, whereas EM is all-or-nothing over the whole document. W-ED measures how far a miss actually is. All three operate on whitespace-delimited words, making them independent of any model-specific subword tokenizer and therefore directly comparable across the Gemma- and Qwen-based LLMAE instantiations, the Mistral-based ICAE, and the BERT-based COSMOS. We report additional similarity and perplexity metrics in App.~\ref{sec:supp_fidelity}, along with formal metric definitions.

\begin{table*}[t]
\centering
\caption{Reconstruction by domain and length (top: up to 512 tokens; bottom: up to 1024). BLEU-4$\uparrow$, exact match (EM)$\uparrow$, word edit distance (W-ED)$\downarrow$; definitions in App.~\ref{sec:supp_fidelity}.}
\label{tab:reconstruction}
\small
\setlength{\tabcolsep}{2.5pt}
\setlength{\aboverulesep}{0pt}
\setlength{\belowrulesep}{0pt}
\renewcommand{\arraystretch}{1.3}
\begin{tabular}{@{} l cc @{\hskip 3pt} ccc @{\hskip 3pt} ccc @{\hskip 3pt} ccc @{}}
\toprule
& & & \multicolumn{3}{c}{\textbf{C4-News-Stratified}} & \multicolumn{3}{c}{\textbf{OpenWebText}} & \multicolumn{3}{c}{\textbf{CreationMMBench}} \\
\cmidrule(lr){4-6} \cmidrule(lr){7-9} \cmidrule(lr){10-12}
\textbf{Method} & \textbf{Latent} & \textbf{\#Par.} & \textbf{BLEU} & \textbf{EM} & \textbf{W-ED} & \textbf{BLEU} & \textbf{EM} & \textbf{W-ED} & \textbf{BLEU} & \textbf{EM} & \textbf{W-ED} \\
\midrule
ICAE & $128{\times}4096$ & 7.46B & 0.831 & 24.2\% & 14.1\% & 0.711 & 22.0\% & 26.5\% & 0.859 & 18.6\% & 10.9\% \\
COSMOS & $512{\times}768$ & 356.5M & 0.783 & 6.0\% & 17.3\% & 0.773 & 3.6\% & 19.4\% & 0.712 & 0.0\% & 26.7\% \\
\multicolumn{12}{@{}l}{\emph{LLMAE (Ours)}} \\
\rowcolor{blue!7} \quad Qwen & $256{\times}896$ & 494M & \textbf{0.999} & \textbf{96.6\%} & 0.12\% & \textbf{0.997} & \textbf{93.8\%} & \textbf{0.17\%} & \textbf{0.998} & \textbf{92.6\%} & \textbf{0.15\%} \\
\rowcolor{blue!7} \quad Gemma & $256{\times}640$ & 278.8M & \textbf{0.999} & 92.6\% & \textbf{0.05\%} & 0.995 & 80.6\% & 0.58\% & 0.994 & 83.9\% & 0.81\% \\
\midrule
ICAE & $256{\times}4096$ & 7.46B & 0.710 & 13.4\% & 25.8\% & 0.602 & 8.0\% & 41.3\% & 0.744 & 8.8\% & 22.4\% \\
\multicolumn{12}{@{}l}{\emph{LLMAE (Ours)}} \\
\rowcolor{blue!7} \quad Qwen & $256{\times}896$ & 494M & \textbf{1.000} & \textbf{97.4\%} & \textbf{0.005\%} & \textbf{0.998} & \textbf{93.2\%} & \textbf{0.17\%} & \textbf{0.999} & \textbf{82.4\%} & \textbf{0.10\%} \\
\rowcolor{blue!7} \quad Gemma & $256{\times}640$ & 278.8M & 0.995 & 70.6\% & 0.26\% & 0.988 & 67.4\% & 1.8\% & 0.991 & 68.8\% & 1.5\% \\
\bottomrule
\end{tabular}
\end{table*}

\subsubsection{Reconstruction Performance}
\label{sec:eval_reconstruction}
We demonstrate near-perfect reconstruction in Table~\ref{tab:reconstruction}, maintaining BLEU-4 scores near or exceeding $0.99$, significantly outperforming both ICAE \citep{ge2023context} and COSMOS \citep{meshchaninov2025cosmos}. Beyond BLEU, LLMAE reproduces the majority of documents \emph{verbatim} (EM up to $97.4\%$ at 1024 tokens) with word edit distances two orders of magnitude below the baselines. We observe that ICAE's reconstruction quality degrades as sequence length increases, with its BLEU score dropping from $0.831$ to $0.710$ on the C4-News-Stratified split. This demonstrates that LLMAE provides a more robust bottleneck than the soft-memory compression used in ICAE, despite being an order of magnitude more parameter-efficient (278.8M vs. 7.46B parameters). Of those 278.8M parameters only 10.7M are trained on Gemma: 737K LoRA adapters, 164K latent token embeddings, and a 9.8M additive codec, so the codec accounts for most of the trainable budget. Notably, we also attempted to train an ICAE-style autoencoder using the lightweight Gemma backbone used in LLMAE, but were unable to achieve high-quality reconstruction (Appendix~\ref{sec:supp_icae}), suggesting that simply applying soft-prompt compression to smaller models is insufficient for high quality reconstruction. Component-wise ablations of the KL regularization and the additive codec are reported in Table~\ref{tab:latent_smoothness}: on Gemma the 9.8M-parameter codec closes much of the remaining gap to perfect reconstruction, while on Qwen the bottleneck alone is already near-lossless but is closed further.

\subsection{Latent Space Distribution}
\label{sec:eval_smoothness}
\begin{table}[h]
\centering
\caption{Latent statistics and reconstruction across the training stages of both backbones.}
\label{tab:latent_smoothness}
\footnotesize
\setlength{\tabcolsep}{3pt}
\setlength{\aboverulesep}{0pt}
\setlength{\belowrulesep}{0pt}
\begin{tabular}{@{}l cccc @{\hskip 8pt} ccc@{}}
\toprule
& \multicolumn{4}{c}{\textbf{Latent distribution}} & \multicolumn{3}{c}{\textbf{Reconstruction}} \\
\cmidrule(lr){2-5} \cmidrule(lr){6-8}
\textbf{Stage} & \textbf{Mean} & \textbf{Std} & \textbf{Max} & \textbf{$D_{\text{KL}}$}\,($\downarrow$) & \textbf{BLEU} & \textbf{EM} & \textbf{W-ED} \\
\midrule
\textit{Gemma-270M}, NTP only & 7.068 & 277.61 & 7296.0 & 38556.0 & 0.978 & 53.6\% & 1.8\% \\
\quad + KL & 0.164 & 8.58 & 1328.0 & 35.36 & 0.967 & 29.8\% & 2.2\% \\
\rowcolor{blue!7} \quad + KL + Codec & -0.038 & 5.08 & 236.0 & \textbf{11.69} & \textbf{0.995} & \textbf{70.6\%} & \textbf{0.26\%} \\
\midrule
\textit{Qwen2.5-0.5B}, NTP only & 0.044 & 2.21 & 24.4 & 2.39 & 0.999 & 92.8\% & 0.07\% \\
\quad + KL & -0.008 & 1.04 & 11.6 & 0.22 & 1.000 & 96.8\% & 0.008\% \\
\rowcolor{blue!7} \quad + KL + Codec & -0.007 & 0.97 & 9.1 & \textbf{0.12} & \textbf{1.000} & \textbf{97.4\%} & \textbf{0.005\%} \\
\bottomrule
\end{tabular}
\end{table}
We evaluate the distribution statistics of the latent space $Z \in \mathbb{R}^{M \times K \times d}$, where $M=500$ is the number of test samples, $K=256$ is the latent token length, and $d$ is the latent dimension ($640$ for Gemma, $896$ for Qwen). We report the latent distributional statistics in Table~\ref{tab:latent_smoothness} by computing global empirical moments mean, standard deviation, and maximum absolute value, pooled across all $n, t, d$ indices. The rows are the three training stages: $\mathcal{L}_{\text{NTP}}$ alone; $\mathcal{L}_{z}$ and $\mathcal{L}_{\text{ref}}$ added with $\mathcal{L}_{z}$ on the raw latent $\mathbf{z}$ (+ KL); and the codec added with $\mathcal{L}_{z}$ moved to $\mathbf{z}_{\text{codec}}$ (+ KL + Codec), whose row reports the statistics of $\mathbf{z}_{\text{codec}}$. We observe that default (NTP-only) training activations exhibit highly irregular distributions on Gemma; KL regularization and the codec normalize these statistics effectively, alongside a net improvement in reconstruction. On Gemma, applying the KL to the raw latent alone trades exact-match fidelity (EM $53.6\% \to 29.8\%$), which the codec recovers and surpasses ($70.6\%$). The two backbones differ sharply in raw latent geometry: Gemma's stage-1 activations are dominated by a few extreme outlier coordinates (Std $277$, Max $7296$) which are normalized effectively utilizing KL regularization and the codec, whereas Qwen's are already near-unit-scale (Std $2.2$).

\begin{table}[h]
    \centering
    \begin{minipage}{0.48\textwidth}
        \centering
        \caption{Reconstruction vs.\ latent layer depth.}
        \label{tab:depth_ablation}
        \small
        \setlength{\tabcolsep}{4pt}
        \setlength{\aboverulesep}{0pt}\setlength{\belowrulesep}{0pt}
        \begin{tabular}{cccc}
            \toprule
            \textbf{Depth} & \textbf{BLEU} ($\uparrow$) & \textbf{EM} ($\uparrow$) & \textbf{W-ED} ($\downarrow$) \\
            \midrule
            \multicolumn{4}{l}{\textit{Gemma-270M (18 layers)}} \\
            2 & 0.001 & 0.0\% & 274\% \\
            5 & 0.589 & 14.8\% & 87.8\% \\
            8 & 0.702 & 15.0\% & 55.6\% \\
            \rowcolor{blue!7} 11 & \textbf{0.978} & \textbf{53.6\%} & \textbf{1.8\%} \\
            14 & 0.222 & 0.0\% & 279\% \\
            \midrule
            \multicolumn{4}{l}{\textit{Qwen2.5-0.5B (24 layers)}} \\
            3 & 0.012 & 0.0\% & 322\% \\
            7 & 0.873 & 13.6\% & 97.7\% \\
            11 & 0.998 & 90.4\% & 0.11\% \\
            \rowcolor{blue!7} 15 & \textbf{0.999} & \textbf{92.8\%} & \textbf{0.07\%} \\
            19 & 0.635 & 9.4\% & 61.3\% \\
            \bottomrule
        \end{tabular}
    \end{minipage}
    \hfill 
    \begin{minipage}{0.48\textwidth}
        \centering
        \caption{Reconstruction vs.\ latent token count.}
        \label{tab:token_ablation}
        \small
        \setlength{\tabcolsep}{4pt}
        \setlength{\aboverulesep}{0pt}\setlength{\belowrulesep}{0pt}
        \begin{tabular}{cccc}
            \toprule
            \textbf{\#Tokens} & \textbf{BLEU} ($\uparrow$) & \textbf{EM} ($\uparrow$) & \textbf{W-ED} ($\downarrow$) \\
            \midrule
            \multicolumn{4}{l}{\textit{Gemma-270M}} \\
            64 & 0.376 & 8.4\% & 163\% \\
            128 & 0.216 & 1.8\% & 434\% \\
            \rowcolor{blue!7} 256 & \textbf{0.978} & 53.6\% & \textbf{1.8\%} \\
            512 & 0.958 & 54.6\% & 7.9\% \\
            1024 & 0.916 & \textbf{70.0\%} & 84.5\% \\
            \midrule
            \multicolumn{4}{l}{\textit{Qwen2.5-0.5B}} \\
            64 & 0.397 & 0.0\% & 260\% \\
            128 & 0.931 & 34.4\% & 5.1\% \\
            \rowcolor{blue!7} 256 & 0.999 & 92.8\% & 0.07\% \\
            512 & \textbf{1.000} & \textbf{98.6\%} & 0.005\% \\
            1024 & \textbf{1.000} & \textbf{98.6\%} & \textbf{0.003\%} \\
            \bottomrule
        \end{tabular}
    \end{minipage}
\end{table}

\subsection{Latent Bottleneck Ablations}
\label{sec:eval_bottleneck_ablation}
The ablations in Tables~\ref{tab:depth_ablation} and~\ref{tab:token_ablation} are run at the NTP-only stage of Table~\ref{tab:latent_smoothness}, without the latent KL or the codec, so that each varied factor is isolated, and use each backbone's own data ordering (curriculum on Gemma, shuffled on Qwen). We systematically vary the intermediate latent depth $D$ and the number of latent tokens $K$ on both backbones. As shown in Table~\ref{tab:depth_ablation}, we observe a clear ``entropy valley'' effect that replicates across backbones: reconstruction peaks at a deep intermediate layer (${\sim}2/3$ depth: layer 11 of 18 on Gemma; layer 15 of 24 on Qwen), whereas shallow and final-stage layers are not effective. We also evaluate the bottleneck width $K$ in Table~\ref{tab:token_ablation}. On Gemma, reconstruction is best at $K{=}256$ and degrades slightly at larger budgets; on Qwen, fidelity instead saturates, with $K{=}512$ reaching $98.6\%$ exact-match and $K{=}1024$ adding nothing further, indicating the usable latent budget is backbone-dependent rather than universal. Low budgets ($K{=}64$) suffer information collapse on both (App.~\ref{sec:supp_length_buckets}). An input-level soft prompt on the same backbone and data reaches BLEU $0.393$ at best (App.~\ref{sec:supp_icae}), and an early-layer bottleneck does not work ($0.001$ at layer 2), whereas moving the read-out to a deep intermediate layer reaches $0.978$. The LoRA-rank sweep is reported in App.~\ref{sec:eval_rank_compression}.

\subsection{Downstream Utility: Latent Diffusion for Image Captioning}
\label{sec:eval_captioning}
\begin{table*}[t]
\centering
\caption{Detailed image captioning. Our model (w/ Gemma) uses only 111M trainable parameters.}
\label{tab:captioning}
\small
\setlength{\tabcolsep}{3.5pt}      
\setlength{\aboverulesep}{0pt}    
\setlength{\belowrulesep}{0pt}    
\renewcommand{\arraystretch}{1.3}  
\begin{tabular}{l rccccc}
\toprule
\textbf{Method} & \textbf{\# Params} & \textbf{Length} & \textbf{VLM} ($\uparrow$) & \textbf{CapArena-Auto} ($\uparrow$) & \textbf{RefCLIP} ($\uparrow$) & \textbf{PPL} ($\downarrow$) \\ \midrule
\rowcolor{gray!5} \multicolumn{7}{l}{\textit{Autoregressive Methods}} \\
Gemini-1.5-Pro & -- & 166.5 & 5.90 & 56.17 & 0.454 & 19.8 \\
LLaVA-1.5-7B & 7.1B & 74.7 & 2.89 & -94.00 & 0.439 & 10.9 \\
Qwen2-VL-2B & 2.2B & 117.4 & 4.64 & -48.67 & 0.454 & 14.0 \\
VLV & 4.2B & 217.6 & 4.30 & -38.17 & 0.452 & 9.6 \\
Share-Captioner & 8.4B & 176.5 & 3.34 & -74.83 & 0.439 & 10.6 \\ 
\midrule
\rowcolor{gray!5} \multicolumn{7}{l}{\textit{Latent Diffusion}: on frozen latent space with frozen SigLIP for image conditioning} \\
\rowcolor{blue!7} w/ Gemma LLMAE & 819M & 164.8 & 3.44 & -62.00 & 0.446 & 26.0 \\
\rowcolor{blue!7} w/ Qwen2.5 LLMAE & 1.1B & 163.6 & 3.06 & -50.50 & 0.449 & 24.0 \\
\rowcolor{blue!2} w/ COSMOS & 943M & 161.5 & 2.24 & -88.33 & 0.440 & 48.5 \\
\bottomrule
\end{tabular}
\end{table*}

We demonstrate downstream utility on detailed image captioning, a multimodal task that remains largely unexplored via latent diffusion despite the paradigm's dominance in text-to-image synthesis \cite{rombach2022high}. Keeping both the visual and linguistic representations frozen (Fig.~\ref{fig:captioning_architecture}), we train only a lightweight diffusion model. By operating on a global latent representation, we can utilize the standard MSE denoising score matching objective:
$$\mathcal{L}_{\text{diff}} = \mathbb{E}_{\mathbf{z}_0,\,t,\,\boldsymbol{\epsilon}}\Big[\big\|\hat{\mathbf{z}}_0(\mathbf{z}_t, t, I) - \mathbf{z}_0\big\|^2\Big],$$
where $\mathbf{z}_0$ is the LLMAE latent of the caption, $\mathbf{z}_t$ its noised version at diffusion time $t$ with noise $\boldsymbol{\epsilon}$, $\hat{\mathbf{z}}_0(\cdot)$ the score network's prediction of the clean latent, and $I$ the SigLIP features of the image. The score network is a 12-layer transformer, directly adapted from \cite{meshchaninov2025cosmos}, details are provided in Appendix~\ref{sec:supp_architecture}. We train one diffusion model per backbone, on the $256{\times}640$ latents of the Gemma LLMAE and on the $256{\times}896$ latents of the Qwen2.5 LLMAE; in both cases the latents stay frozen and are decoded into text by the frozen LLMAE decoder during inference (App.~\ref{sec:supp_latent_determines_text}). For image conditioning, a linear layer projects SigLIP \cite{zhai2023sigmoid} patch tokens to the score network width, fed via cross-attention between the score network intermediate layers and these projected visual features.

\textbf{Training data and evaluation details:} We use the VLV-6M dataset \cite{zhang2025vision} to train our latent diffusion model. This is a 6 million subset of LAION-Aesthetic with captions generated by Gemini-2-Flash. To assess the captioning performance, we evaluate our model on the \textit{CapArena-Auto} benchmark \cite{cheng2025caparena}. CapArena is a human-preference benchmark for detailed captioning to establish a preference-based Elo ranking. CapArena-Auto serves as an automated proxy for this protocol; it utilizes a set of 600 test images from the DOCCI dataset \cite{onoe2024docci} and employs GPT-4o as a pairwise judge. This automated protocol was shown to achieve a 94.3\% correlation with human model rankings while significantly reducing evaluation overhead \cite{cheng2025caparena}. We include four metrics to capture distinct measurements on the same 600-image test set used by CapArena:
\begin{itemize}[leftmargin=*,noitemsep]
    \item \textbf{CapArena-Auto Score:} The mean normalized win rate against 3 reference VLM captioners.
    \item \textbf{VLM Judge (0-6):} We adopt the system prompt with a 7-point rubric from \cite{zhang2025vision} (Fig.~\ref{fig:vlm_judge_prompt}). This system comprehensively evaluates the coverage of image elements, spatial layout consistency, and the absence of hallucinations. We use \texttt{gpt-4o-2024-08-06} as the VLM judge.
    \item \textbf{RefCLIPScore:} The harmonic mean of CLIP similarities between (caption, image) and (caption, reference) using \texttt{OpenCLIP ViT-B/32}, capturing semantic alignment \cite{hessel2021clipscore}.
    \item \textbf{Perplexity (PPL):} of generated captions, using GPT-2-Large, to measure the linguistic fluency.
\end{itemize}

We evaluate against three representative models from the CapArena leaderboard: Gemini-1.5-Pro \citep{team2024gemini} is the best performer, LLaVA-1.5-7B \citep{liu2024improved} as a poor performer, and Qwen2-VL-2B \citep{wang2024qwen2} as a relatively lightweight VLM. We also compare against dedicated autoregressive captioning models, Share-Captioner \cite{chen2024sharegpt4v} and VLV \cite{zhang2025vision}. Notably, all models fine-tune the underlying LLM backbone, whereas our approach maintains a completely frozen LLMAE autoencoder and a frozen SigLIP image encoder.

With the Gemma LLMAE, our lightweight latent diffusion model achieves a VLM Judge score of 3.44, outperforming both the 7.1B LLaVA-1.5 (2.89) and the specialized Share-Captioner (3.34), while remaining behind the stronger VLM captioners (Qwen2-VL-2B, VLV, and Gemini-1.5-Pro). Generative reliability is high: no caption is marked as non-informative (score 0) and 87.5\% achieve a score of 2 or higher, which designates a caption as ``informative'' with valid semantic grounding under the VLV rubric. Substituting the Qwen2.5 LLMAE under the same protocol improves the pairwise CapArena-Auto score, RefCLIPScore and perplexity, but lowers the VLM Judge score ($81.0\%$ of captions at 2 or higher, four scored 0), so the two backbones trade off between the pairwise and the rubric judge. Substituting our autoencoder with COSMOS \cite{meshchaninov2025cosmos}, while maintaining the identical diffusion architecture and parameters, which itself is adapted from their original work, results in a significant performance degradation, confirming that the high-fidelity LLMAE autoencoder is better suited for this task. A representative caption scored $6/6$ by the VLM judge is shown in Fig.~\ref{fig:captioning_success} in Appendix~\ref{sec:supp_caption_quality}, alongside a failure case; a granular analysis of the failure modes there reveals that the model demonstrates high fidelity in describing aesthetic scenes and atmospheric nuances, reflecting the training data, but fails on text and symbols in the image.

\section{Limitations}
Despite the strong performance and efficiency of LLMAE, limitations remain. First, due to using a fine-tuned LLM, LLMAE requires autoregressive sampling in decoding that causes sequential latency.  Encoding is still parallelizable. Because the masking protocol respects causal ordering in each layer, the framework is compatible with standard KV caching \cite{pope2023efficiently}. Second, we did not perform any experiments for sequence lengths longer than 1024 tokens of text, roughly 4.6K characters or 760 words. Finally, we observe a fundamental constraint involving the balance of autoencoding fidelity, sequence-length reduction, and latent smoothness. By prioritizing high-resolution, near-lossless reconstruction, LLMAE creates a dense latent manifold (App.~\ref{sec:supp_latent_interpolation}) that may require a strong conditional signal for generative modeling. Future research may help address some of these concerns without sacrificing the reconstruction quality.
\section{Conclusion}
We introduced LLMAE, a framework for repurposing pre-trained decoder-only language models as high-fidelity continuous text autoencoders with a fixed-length latent. What makes this work is \emph{where} the attention-mask bottleneck is placed: activations read from a deep intermediate layer reconstruct documents of up to 1024 tokens almost verbatim. We utilize low-rank adaptation to perform parameter-efficient fine-tuning. The depth optimum sits near two-thirds of the stack on both backbones (layer 11 of 18 on Gemma, 15 of 24 on Qwen), an empirical confirmation of the ``entropy valley'' that gives a principled basis for extracting dense latents for text reconstruction. Furthermore, our results in latent text diffusion for detailed image captioning validate the downstream utility of this representation. Ultimately, LLMAE offers a parameter-efficient text autoencoder for grounding continuous generative models in the rich priors of LLMs, enabling flexible architectures for multimodal tasks and contributing to more structurally aware, globally consistent text generation.

\clearpage
\bibliography{references}
\bibliographystyle{unsrt}

\clearpage
\appendix
\section*{Appendix: Repurposing Pre-trained LLMs as High Fidelity Continuous Text Autoencoders}
\subsection*{Outline}

\begin{description}[leftmargin=1.2em, style=nextline, font=\normalfont]
    \item[\ref{sec:supp_caption_quality}: \textbf{Captioning Model Quality Analysis}] 
    A detailed qualitative analysis, identifying high aesthetic fidelity but critical failures in symbolic grounding and OCR.
    
    \item[\ref{sec:supp_latent_determines_text}: \textbf{Captioning: Global Content is Determined in Latent Space}]
    Stochastic decodes of one latent versus decodes of independently sampled latents.

    \item[\ref{sec:sup_attention_masking_strategy}: \textbf{Attention Masking Strategy}]
    Detailed breakdown of the attention visibility matrix for LLMAE.
    
    \item[\ref{sec:supp_latent_smoothness}: \textbf{LLMAE Latent Smoothness}] 
    Latent robustness to Gaussian noise perturbations and $D_{\text{KL}}$ calculation equation.
    
    \item[\ref{sec:supp_latent_interpolation}: \textbf{LLMAE Latent Interpolation}] 
    Linear interpolation examples to observe phrase-level recombination and syntactic preservation.

    \item[\ref{sec:supp_curriculum_ablation}: \textbf{Data Ordering Ablation}]
    The length curriculum versus shuffled training data on both backbones.

    \item[\ref{sec:eval_rank_compression}: \textbf{Adapter Rank and Compressing $K$ further}]
    The LoRA-rank sweep on both backbones, and what compressing a trained $K{=}256$ latent to $64$ tokens shows.
    
    \item[\ref{sec:supp_length_buckets}: \textbf{Reconstruction by Input Length}]
    The latent-budget sweep broken out by input length, showing where each $K$ succeeds and fails.

    \item[\ref{sec:supp_icae}: \textbf{ICAE-style Autoencoder with Gemma-270M}]
    Comparative baseline ablation when using the ICAE architecture instead of LLMAE.

    \item[\ref{sec:supp_exact_fidelity}: \textbf{Exact Reconstruction Fidelity Metrics}]
    Exact-match and edit-distance metric definitions with a per-category error breakdown.

    \item[\ref{sec:supp_architecture}: \textbf{Method Details}]
    Detailed specifications for LLMAE and the latent diffusion model for captioning.
\end{description}

\section{Captioning Model Quality Analysis}
\label{sec:supp_caption_quality}

\begin{figure}[h]
    \centering
    \begin{tikzpicture}
        \begin{axis}[
            ybar,
            bar width=18pt, 
            ylabel={Count},
            symbolic x coords={1, 2, 3, 4, 5, 6},
            xtick=data,
            nodes near coords,
            point meta=explicit symbolic,
            every node near coord/.append style={
                font=\footnotesize, 
                rotate=90, 
                anchor=west,
                yshift=0pt,
            },
            ymin=0, ymax=330, 
            enlarge x limits=0.15,
            axis lines*=left,
            ymajorgrids=true,
            grid style=dashed,
            width=0.75\textwidth, 
            height=6cm            
        ]
            \addplot[fill=cyan!60, draw=black] coordinates {
                (1, 75) [75 (12.5\%)]
                (2, 255) [255 (42.5\%)]
                (3, 3) [3 (0.5\%)]
                (4, 38) [38 (6.3\%)]
                (5, 55) [55 (9.2\%)]
                (6, 174) [174 (29.0\%)]
            };
        \end{axis}
    \end{tikzpicture}
    \caption{Distribution of VLM-Judge scores across the evaluation dataset.}
    \label{fig:score_histogram}
\end{figure}

\begin{figure}[h]
    \begin{tcolorbox}[
        colback=gray!8,
        colframe=gray!40,
        arc=4pt,
        boxrule=0.5pt,
        width=\linewidth,
        left=10pt, right=10pt, top=10pt, bottom=10pt
    ]
        \small
        \textbf{VLM Score Judge Prompt:} 
        \vspace{0.5em}
        
        Your role is to serve as an impartial and objective evaluator of an image caption generated by a Large Multimodal Model (LMM). Based on the single image input, assess the caption on three main criteria:

        \begin{enumerate}[leftmargin=*, noitemsep, topsep=4pt]
            \item Coverage of image elements – how well the caption mentions the salient objects, their attributes, actions, and contextual details.
            \item Absence of hallucinations – the caption must not invent objects, attributes, counts, spatial relations, or other details not present or implied by the image.
            \item Object spatial layout consistency – whether spatial relationships (left/right, above/below, front/behind, center, background/foreground) are described accurately.
            \begin{itemize}[label=$\bullet$, noitemsep, topsep=2pt]
                \item Any incorrect or invented spatial relation is a hallucination.
                \item Omitting an obvious spatial relation reduces coverage.
                \item Stating a relation that is ambiguous or uncertain in the image is also a hallucination.

            \end{itemize}
        \end{enumerate}

        \vspace{0.8em}
        \upshape Evaluation protocol:\\
        Start with a brief explanation of your evaluation process. Then assign one rating using the scale below. Output only the rating number—no extra text, symbols, or commentary.

        \vspace{0.5em}
        \begin{description}[leftmargin=1.5em, noitemsep, font=\bfseries\upshape, style=nextline]
            \item[6] Comprehensive coverage, correct spatial layout, no hallucinations
            \item[5] Very informative, correct spatial layout, no hallucinations, minor omissions
            \item[4] Moderate coverage, correct spatial layout, no hallucinations, several omissions
            \item[3] Limited coverage, minimal spatial detail, no hallucinations
            \item[2] Informative but contains at least one hallucination (object or spatial)
            \item[1] Limited coverage and at least one hallucination (object or spatial)
            \item[0] Not informative and/or multiple hallucinations
        \end{description}
    \end{tcolorbox}
    \caption{System prompt used for the VLM Score Judge evaluation, adapted exactly from \cite{zhang2025vision}.}
    \label{fig:vlm_judge_prompt}
\end{figure}
\begin{figure}[h]
    \begin{tcolorbox}[
        colback=gray!8,       
        colframe=gray!40,     
        arc=4pt,              
        boxrule=0.5pt,        
        width=\linewidth,     
        left=8pt, right=8pt, top=8pt, bottom=8pt 
    ]
        \begin{wrapfigure}{l}{0.35\textwidth}
            \vspace{-15pt} 
            \centering
            \includegraphics[width=\linewidth]{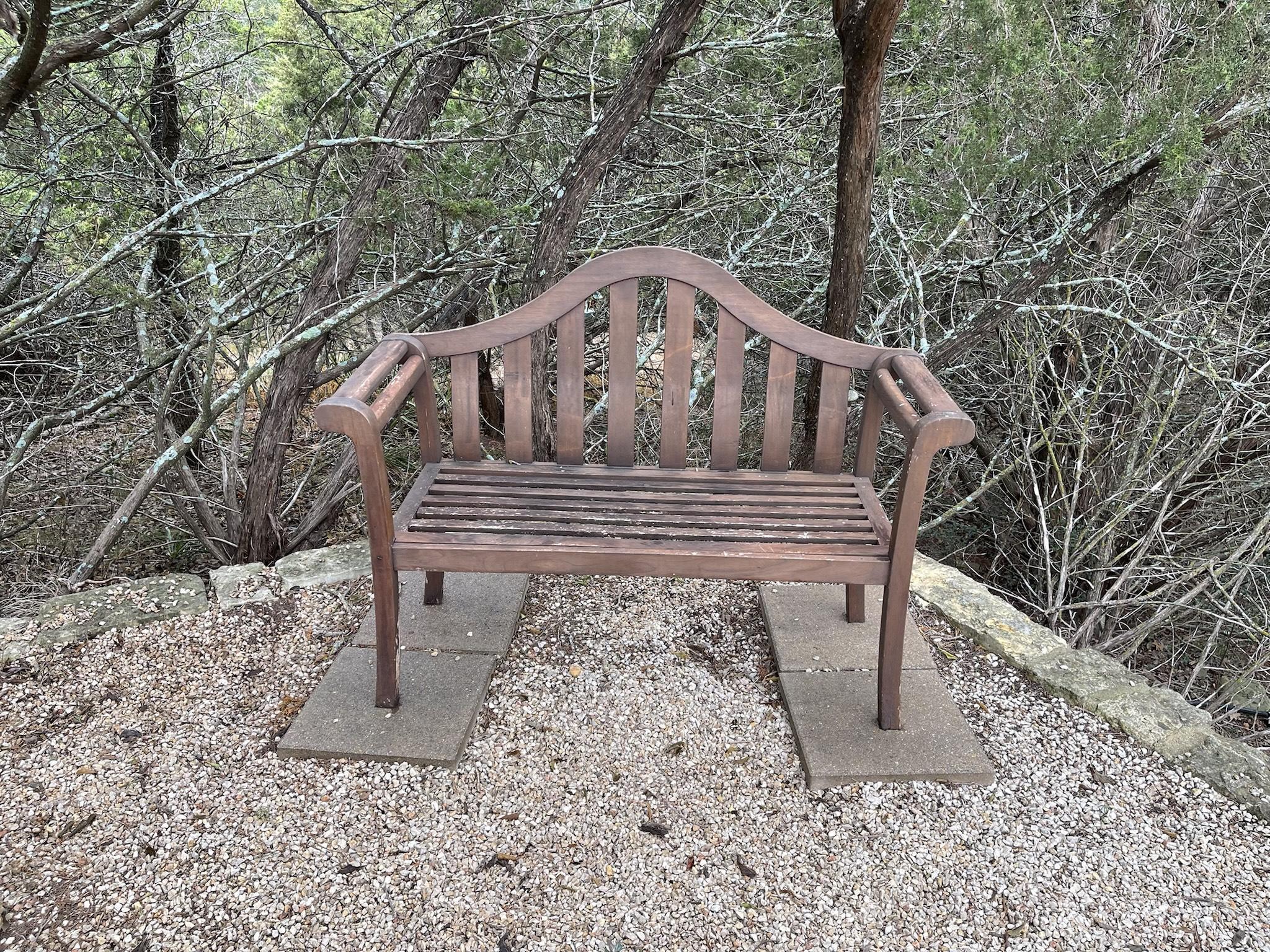}
            \vspace{-25pt} 
        \end{wrapfigure}
        
        \textbf{VLM Judge Score:} 6/6 \par\vspace{0.4em}
        \textbf{Caption:} \textit{A full shot captures a wooden bench positioned amidst a natural, forest-like setting. The wooden bench, the focal point of the image, prominently features a curved back, armrests, and vertical slats that enhance its classic design. It rests on a rectangular stone platform, flanked by two stout, light-colored bases. The ground surrounding the bench or platform is covered in small, light gray gravel, creating a neutral and textured surface. The background consists of dense foliage, predominantly leafless trees and bushes, indicating a natural and serene environment. These branches reach towards the sky, lending a sense of depth and enclosure to the scene.\\\\The lighting appears soft and diffused, casting gentle shadows across the scene, suggesting an overcast day. The dominant colors are muted greens, greys, and a muted touch of somber tones. The texture varies from the smooth, weathered wood of the bench to the rough stone and the smooth bark of the trees and branches.\\\\There are no people or animals visible in the frame. The overall mood is peaceful, evoking a sense of quiet contemplation within the natural surroundings. The color palette is dominated by shades of brown, beige, and grey, which contribute to the earthy and muted aesthetic of the scene.}
    \end{tcolorbox}
    \caption{\textbf{Latent Diffusion with LLMAE.} Qualitative demonstration of detailed captioning ability.}
    \label{fig:captioning_success}
\end{figure}
\begin{figure}[t]
    \begin{tcolorbox}[
        colback=gray!8,       
        colframe=gray!40,     
        arc=4pt,              
        boxrule=0.5pt,        
        width=\linewidth,     
        left=8pt, right=8pt, top=8pt, bottom=8pt 
    ]
        \begin{wrapfigure}{l}{0.35\textwidth}
            \vspace{-15pt} 
            \centering
            \includegraphics[width=\linewidth]{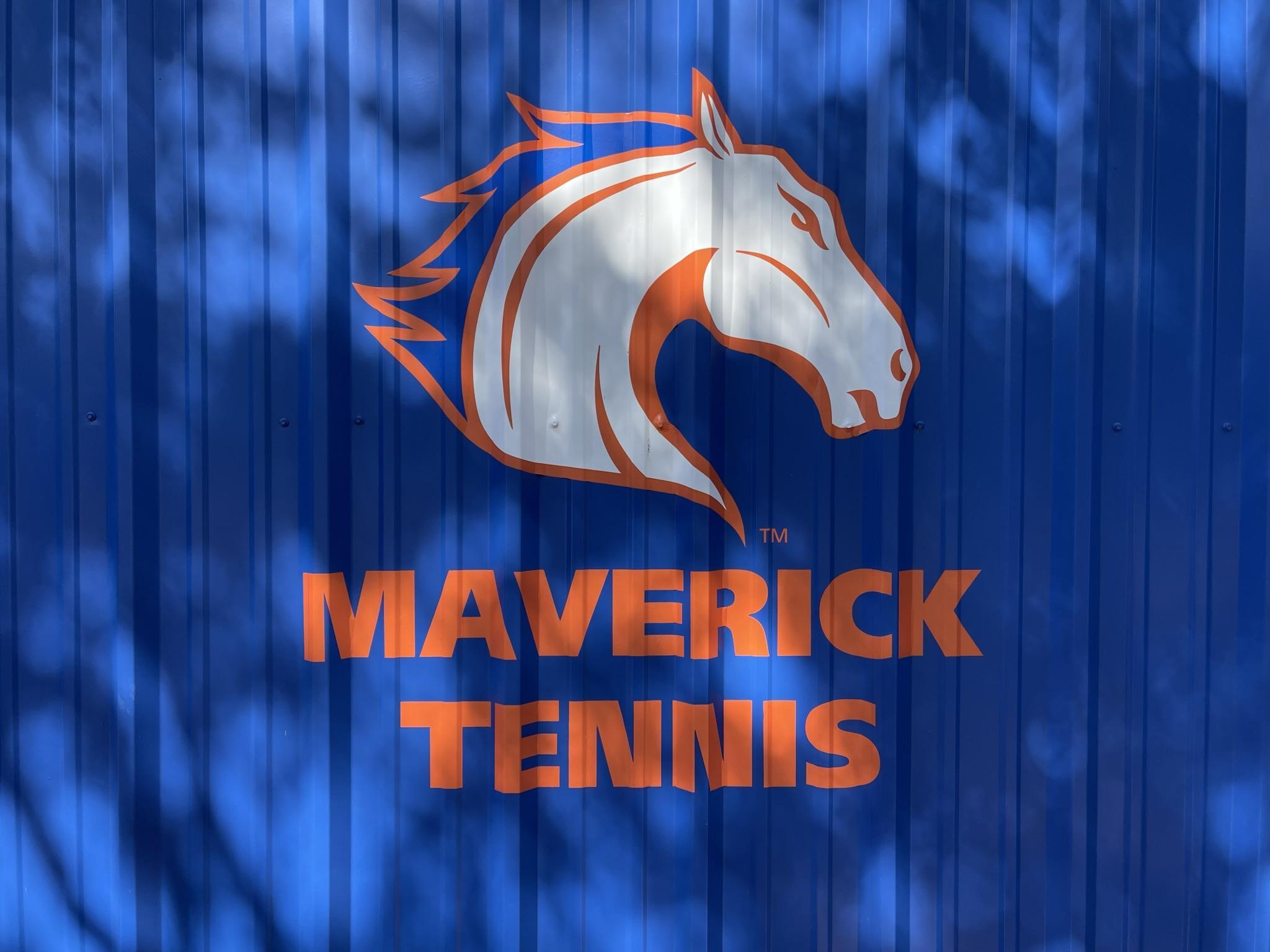}
            \vspace{-25pt} 
        \end{wrapfigure}
        
        \textbf{Caption:} \textit{``The image is a close-up shot of Convention Tennis Tennis against a blue corrugated backdrop. The prominent lettering ``CorrespondingON'' features a stylized depiction of a horse head with a mane and tail, both outlined in bright orange with a subtle gradient highlight. The horse is rendered in profile, facing right, with its head raised, giving a slight frown. The backdrop is a vibrant blue, vertically ribbed metal, creating vertical shadows that suggest it is daytime. Below the horse's emblem, the words ``VICK'' are printed in a classic, bold orange font. Below the horse, the words ``Correspondence TNINS'' are written in the same orange type, slightly smaller font. The overall lighting is bright and even, emphasizing the contrast between the blue metal backdrop and the vivid colors. The mood conveyed is sleek and bold, appropriate for school team visual's nod to tennis players.''}  \par\vspace{0.4em}
        \textbf{VLM Judge Score:} 1/6 \par\vspace{0.4em}
        \textbf{VLM Judge Explanation:} The caption contains several inaccuracies and hallucinations. It incorrectly describes the text as ``Convention Tennis Tennis'' and ``CorrespondingON,'' while the actual text is ``MAVERICK TENNIS.'' It also mentions ``Correspondence TNINS'' and ``VICK,'' which are not present. The description of the horse's mane and tail is inaccurate, as the image only shows the horse's head. The spatial layout and lighting description are mostly correct, but the caption's errors significantly impact its accuracy.
    \end{tcolorbox}
    \caption{Qualitative demonstration of the captioning model's lack of grounding.}
    \label{fig:captioning_failure}
\end{figure}

We provide the distribution of VLM Judge Scores of our captioning model on the CapArena test set in Fig.~\ref{fig:score_histogram}. Fig.~\ref{fig:captioning_success} shows a caption scored $6/6$ by the judge, and Fig.~\ref{fig:captioning_failure} one scored $1/6$.

\subsection{Categorical Performance Analysis}
To analyze the failure modes and performance benchmarks of our captioning model, we utilized an LLM Judge to aggregate and categorize the qualitative feedback and numerical scores from the full set of per-image VLM Judge explanations generated by GPT-4o. The following LLM-generated summary highlights a significant performance gap between aesthetic scene interpretation and symbolic grounding:

\begin{itemize}
    \item \textbf{Environmental and Aesthetic Fidelity (Primary Strength):} The model demonstrates high fidelity in describing natural landscapes and organic textures (Scores: 5-6). It utilizes an advanced vocabulary to describe environmental nuances such as moss-covered surfaces, leaf translucency, and specific floral species. Furthermore, it excels in atmospheric reasoning, correctly identifying lighting conditions such as dappled sunlight, diffused overcast light, and artificial nighttime illumination.
    
    \item \textbf{Compositional and Spatial Awareness:} There is strong evidence of high-level compositional understanding. The model consistently identifies camera perspectives (e.g., macro, low-angle, aerial) and correctly interprets the spatial relationship between foreground framing elements and background subjects. This suggests an effective global awareness of the scene's visual hierarchy.
    
    \item \textbf{Symbolic and OCR Grounding (Critical Failure):} Systematic failure occurs in tasks requiring the decoding of textual or symbolic data (Scores: 1-2). The model frequently exhibits ``symbolic hallucination,'' where it confidently generates plausible but entirely incorrect text for signs, murals, and labels (e.g., misreading \textit{MAVERICK TENNIS} as \textit{Convention Tennis Tennis} in Fig.~\ref{fig:captioning_failure}). This indicates a lack of robust low-level OCR integration.
    
    \item \textbf{Quantitative Reliability and Individuation:} The model struggles with discrete object counting and instance individuation. It consistently misreports quantities (e.g., reporting ``ten birds'' for significantly larger flocks) and occasionally hallucinates secondary objects to ``fill'' the semantic space of a scene.
    
    \item \textbf{Domain-Specific Technical Accuracy:} Performance degrades in industrial or specialized contexts due to ``false specificity.'' While the model correctly identifies general categories (e.g., ``vehicle''), it often hallucinates incorrect brand names, model numbers, or mechanical components (e.g., attributing multiple turrets to single-turret tanks), suggesting a limit to its fine-grained classification capabilities.
    
    \item \textbf{Linguistic Consistency:} A recurring failure mode involves ``lexical stuttering'' or repetitive token generation within long-form descriptions (e.g., \textit{``fleece and fleece''} or \textit{``low-angle, low-angle''}). This suggests periodic breakdowns in the model's self-attention mechanism during the autoregressive generation of complex captions.
\end{itemize}

\section{Captioning: Global Content is Determined in Latent Space}
\label{sec:supp_latent_determines_text}

Sec.~\ref{sec:eval_captioning} decodes each sampled latent with the frozen LLMAE decoder, treating the decode as a rendering step over a latent that already fixes the content. We test this directly on 24 evaluation images, using the Gemma captioning model under the sampling protocol of Table~\ref{tab:captioning}. For each image we compare two sources of randomness, each contributing five samples and therefore ten pairwise comparisons. In the first we sample a single latent and decode it five times with stochastic sampling (temperature $1.0$), varying only the decode seed. In the second we sample five independent latents for the same image and decode each once, holding the decode seed fixed so that only the latent varies. Differences between two captions are measured as word-level edit distance normalized by the longer caption ($0$ identical, $1$ entirely different). If global content were decided during decoding, the first condition would diverge; if it is determined by the latent, only the second should.

\begin{table}[h]
    \centering
    \caption{Caption variation by source of randomness, over 24 evaluation images.}
    \label{tab:latent_determines_numbers}
    \small
    \begin{tabular}{llcccc}
        \toprule
        \textbf{Varying} & \textbf{Held fixed} & \textbf{Mean} & \textbf{Median} & \textbf{Per-image range} & \textbf{Identical pairs} \\
        \midrule
        Decode seed & Latent      & 0.016 & 0.010 & 0.000-0.063 & 29.2\% \\
        Latent      & Decode seed & \textbf{0.829} & \textbf{0.823} & 0.746-0.902 & 0.0\% \\
        \bottomrule
    \end{tabular}
\end{table}

Table~\ref{tab:latent_determines_numbers} reports the result. Varying the decode seed changes almost nothing: a mean edit distance of $0.016$, with $29.2\%$ of decode pairs byte-identical and all five decodes identical on 5 of the 24 images. Varying the latent changes nearly everything: a mean of $0.829$, roughly $50\times$ larger, with no identical pair anywhere in the condition. The separation holds on every image individually, as the per-image ranges do not overlap (0.000-0.063 against 0.746-0.902), so it is not an artifact of averaging. Decoding stochastically rather than greedily from the same latent is equally inconsequential (mean $0.011$). Table~\ref{tab:latent_determines_examples} shows both conditions for one image: two decodes of one latent agree on 175 of 176 words, differing only in \emph{toward} versus \emph{towards}, whereas two independent latents produce captions that remain grounded in the same scene --- the squirrel, the tile surface, the white pillar, the surrounding greenery --- but differ in which details are selected, where they appear, and the phrasing throughout. Content and structure are therefore properties of the latent, not of the decoding process, which complements the near-exact reconstruction of Table~\ref{tab:reconstruction} and the noise robustness of Table~\ref{tab:noise_ablation}.

\begin{table}[h]
    \centering
    \caption{Both conditions for one evaluation image. Bold marks the only difference (top) and the points of divergence (bottom).}
    \label{tab:latent_determines_examples}
    \small
    \begin{tabular}{@{}lp{0.86\linewidth}@{}}
        \toprule
        \multicolumn{2}{@{}l}{\emph{Same latent}, two stochastic decodes --- 175 of 176 words identical (word ED $0.006$)} \\
        \midrule
        Decode 1 & \ldots the primary subject of the image, is facing forward with its body turned slightly \textbf{toward} the camera. It has a long tail and a bushy\ldots \\
        Decode 2 & \ldots the primary subject of the image, is facing forward with its body turned slightly \textbf{towards} the camera. It has a long tail and a bushy\ldots \\
        \midrule
        \multicolumn{2}{@{}l}{\emph{Two independent latents}, same image (word ED $0.804$)} \\
        \midrule
        Latent 1 & A medium shot captures a squirrel standing on a \textbf{tile surface near a white pillar, with a glimpse of a green lawn and mulch}. The squirrel, \textbf{the primary subject of the image, is facing forward with its body turned slightly toward the camera. It has a long tail and a bushy tail}\ldots \\
        Latent 2 & A medium shot captures a squirrel standing on a \textbf{tile patio, with a white column and support element in the outdoor setting}. The squirrel, \textbf{positioned towards the lower center of the frame, has a bushy gray fur and tan tail}\ldots \\
        \bottomrule
    \end{tabular}
\end{table}

\section{Attention Masking Strategy}
\label{sec:sup_attention_masking_strategy}
Table~\ref{tab:masking} provides a detailed breakdown of the attention masking strategy for LLMAE.
\begin{table}[t]
\centering
\caption{Attention masking strategy managing visibility between the encoding Input, Embed tokens, and decoder Output targets. Rows denote queries; columns denote keys. For a query token, \textbf{Visible} keys are subject to causal masking to maintain autoregressive consistency, while \textbf{Masked} keys prevent information flow. Layer 0 serves as a unique interface where the Output tokens are permitted to attend to the Embed tokens, effectively providing a global initialization for the reconstruction segment.}
\label{tab:masking}
\small
\begin{tabular}{lcccc}
\toprule
\textbf{Block (Layers)} & \textbf{Query} & \textbf{Key: Input} & \textbf{Key: Embed} & \textbf{Key: Output} \\
\midrule
\textbf{Input Layer} & Output & Masked & Visible & Visible \\
(Layer 0) & Embed & Visible & Visible & Masked \\
\midrule
\textcolor{encodingblue}{\textbf{Encoder Block}} & Output & Masked & Masked & Visible \\
(Layers 1 to $D$) & Embed & Visible & Visible & Masked \\
\midrule
\textcolor{decodingorange}{\textbf{Decoder Block}} & Output & Masked & Visible & Visible \\
(Layers $D+1$ to $L-1$) & Embed & Masked & Visible & Masked \\
\bottomrule
\end{tabular}
\end{table}

\section{LLMAE Latent Smoothness}
\label{sec:supp_latent_smoothness}

To measure prior alignment in Table~\ref{tab:latent_smoothness}, we treat each coordinate $(t, d)$ as a marginal distribution $q_{td}$ and compute its empirical statistics over the $M$ test samples, $\hat{\mu}_{td} = \frac{1}{M} \sum_{n=1}^{M} z_{n,t,d}$ and $\hat{\sigma}^2_{td} = \frac{1}{M} \sum_{n=1}^{M} (z_{n,t,d} - \hat{\mu}_{td})^2$. Same as the KL regularization loss calculation, we compute $D_{\text{KL}}$ analytically using Eq.~\ref{eq:kl_divergence}.

\begin{equation}
    D_{\text{KL}}(q_{td} \parallel \mathcal{N}(0, 1)) = \frac{1}{2} \left( \hat{\sigma}^2_{td} + \hat{\mu}^2_{td} - 1 - \ln \hat{\sigma}^2_{td} \right)
    \label{eq:kl_divergence}
\end{equation}
During training the same divergence serves as $\mathcal{L}_{z}$, with $\hat{\mu}_{td}$ and $\hat{\sigma}^2_{td}$ estimated over the samples of the current batch and Eq.~\ref{eq:kl_divergence} averaged over all $K \times d$ coordinates $(t, d)$.

We further investigate the robustness of LLMAE autoencoding through noise perturbations: we apply additive i.i.d. Gaussian noise to every scalar dimension: $z'_{n,t,d} = z_{n,t,d} + \sigma \epsilon_{n,t,d}$, where $\epsilon_{n,t,d} \sim \mathcal{N}(0, 1)$. As shown in Table~\ref{tab:noise_ablation}, the generative prior of Gemma 3 effectively denoises the latent signal, maintaining high performance up to $\sigma = 0.5$.
\begin{table}[h]
    \centering
    \caption{Quality w.r.t. noise augmentation (Gemma, 1024-token setting).}
    \label{tab:noise_ablation}
    \begin{tabular}{cccc}
        \toprule
        \textbf{Noise $\sigma$} & \textbf{BLEU} ($\uparrow$) & \textbf{PPL} ($\downarrow$) & \textbf{BERT} ($\uparrow$) \\ 
        \midrule
        0 & 0.995 & 24.4 & 0.999 \\
        0.1 & 0.995 & 24.5 & 0.999 \\
        0.25 & 0.991 & 24.7 & 0.998 \\
        0.5 & 0.927 & 31.2 & 0.993 \\
        \bottomrule
    \end{tabular}
\end{table}

We further probe error tolerance directly across autoencoders by perturbing
encoded latents in $z$-space at larger magnitudes and measuring the word
similarity of the decoded text against each model's own clean ($\sigma{=}0$)
decode (Table~\ref{tab:noise_tolerance}). Within the stage-1 (NTP-only) family,
tolerance rises monotonically with LoRA rank. We note that the codec contracts
the latent scale (Table~\ref{tab:latent_smoothness}), so a fixed absolute
$\sigma$ corresponds to a larger \emph{relative} perturbation for the codec
latent; the values below are therefore comparable within, but not directly
across, latent scales.

\begin{table}[h]
    \centering
    \caption{Latent error tolerance: word similarity of the decoded text vs.\ each model's clean decode, under i.i.d.\ Gaussian $z$-space noise. (Qwen unless noted.)}
    \label{tab:noise_tolerance}
    \small
    \begin{tabular}{lccc}
        \toprule
        \textbf{Autoencoder} & \textbf{$\sigma{=}1.0$} & \textbf{$\sigma{=}1.5$} & \textbf{$\sigma{=}2.0$} \\
        \midrule
        NTP only, $r{=}8$  & 0.9952 & 0.9337 & 0.6364 \\
        NTP only, $r{=}16$ & 0.9967 & 0.9839 & 0.8519 \\
        NTP only, $r{=}32$ & \textbf{1.0000} & \textbf{0.9975} & \textbf{0.9730} \\
        KL + Codec         & 0.9994 & 0.8528 & 0.2627 \\
        \midrule
        Gemma KL + Codec   & 0.9947 & 0.9426 & 0.5731 \\
        \bottomrule
    \end{tabular}
\end{table}

\section{LLMAE Latent Interpolation}
\label{sec:supp_latent_interpolation}
To better understand the geometry of the learned text autoencoder, we perform a simple latent interpolation experiment. Given two input texts $x_A$ and $x_B$, we encode them with the LLMAE encoder $E(\cdot)$ into latent representations
\[
z_A = E(x_A), \qquad z_B = E(x_B),
\]
and decode a midpoint linear interpolant:
\[
z_{\text{mid}} = \frac{z_A + z_B}{2}
\]
This experiment is not intended to show that Euclidean interpolation corresponds to semantic interpolation. Rather, it probes whether intermediate latent states remain decodable and what kinds of structure are preserved between distant texts. 
We observe a desirable property of the autoencoder that the interpolated latents often decode into recognizable, locally coherent text rather than complete garbage. However, the decoded midpoint is usually not a meaningful semantic average. Instead, it frequently preserves the global syntactic frame of one or both endpoints while swapping entities, modifiers, actions, or causal clauses. This suggests that the latent space supports high-resolution phrase-level recombination, even though linear paths through it do not reliably preserve semantic consistency. We provide examples in Table~\ref{tab:text-ae-interpolation}.
\begin{table*}[t]
\centering
\caption{
Linear interpolation in the text autoencoder latent space. Midpoints generally remain fluent and often preserve sentence-level structure, but they do not correspond to clean semantic averages. Instead, they exhibit phrase-level recombination: sometimes producing coherent hybrids and sometimes producing binding errors.
}
\small
\begin{tabular}{p{0.15\linewidth} p{0.24\linewidth} p{0.24\linewidth} p{0.24\linewidth}}
\toprule
\textbf{Pair} & \textbf{$x_A$} & \textbf{$x_B$} & \textbf{Decoded midpoint} \\
\midrule
Entity and object swap &
A tired tabby cat stretched across the warm windowsill while rain tapped against the glass and a half-finished cup of tea cooled beside an open novel. &
A tired golden retriever stretched across the warm windowsill while rain tapped against the glass and a half-finished cup of coffee cooled beside an open newspaper. &
A tired \textbf{golden cat} stretched across the warm windowsill while rain tapped against the glass and a half-finished cup of \textbf{coffee} cooled beside an open \textbf{novel}. \\
\midrule
Action and scene swap &
The young violinist stood beneath the theater lights, closed her eyes, and played a slow melody while the audience listened in complete silence. &
The young magician stood beneath the theater lights, raised his hands, and released a cloud of silver confetti while the audience cheered in surprise. &
The young \textbf{violinist} stood beneath the theater lights, \textbf{raised his hands}, and released a \textbf{slow melody while confetti} while the audience \textbf{complete silence surprise}. \\
\midrule
Scientific domain swap &
During the biology lab, the students carefully observed yeast cells under a microscope, recorded changes in their growth, and argued that temperature was affecting the culture. &
During the astronomy workshop, the students carefully observed distant galaxies through a telescope, recorded changes in their brightness, and argued that dust was affecting the measurement. &
During the \textbf{astronomy lab}, the students carefully observed \textbf{distant galaxies} under a \textbf{telescope}, recorded changes in their \textbf{brightness}, and argued that \textbf{temperature} was affecting the \textbf{measurement}. \\
\midrule
Opposing policy claims &
The city should expand protected bike lanes because safer streets reduce traffic injuries, encourage commuting without cars, and make neighborhoods easier to navigate. &
The city should remove protected bike lanes because narrower roads slow emergency vehicles, frustrate local businesses, and make neighborhoods harder to navigate. &
The city should \textbf{expand} protected bike lanes because \textbf{safer roads slow emergency injuries}, \textbf{frustrate commuting without cars}, make make \textbf{harder to navigate}. \\
\midrule
Different domains, similar tone &
After years of drought, the farmer knelt in the cracked field, pressed a handful of dry soil between his fingers, and wondered whether the next storm would arrive in time. &
After years of training, the astronaut floated beside the station window, watched Earth turn silently below her, and wondered whether the next signal would arrive in time. &
After years of \textbf{training}, the \textbf{astronaut floated beside the station window}, \textbf{pressed Earth turn of dry her between his fingers}, the next \textbf{signal} would arrive in time arrive in time. \\
\midrule
Completely different texts &
A legal memorandum explains that the contract cannot be enforced because the signature was forged, the deadline had expired, and the witness testimony contradicted the filing. &
A fairy-tale paragraph describes a child following a blue firefly through an enchanted forest, discovering a hidden door in an oak tree, and hearing music from an invisible castle. &
A legal memorandum explains \textbf{paragraph describes} contract cannot be enforced because \textbf{firefly through forged enchanted forest}, discovering a \textbf{hidden door} in witness testimony contradicted the filing. \textbf{music from an invisible castle}. \\
\bottomrule
\end{tabular}
\label{tab:text-ae-interpolation}
\end{table*}

\section{Data Ordering Ablation}
\label{sec:supp_curriculum_ablation}

\paragraph{Data ordering.}
Table~\ref{tab:ordering} shows that the short-to-long length curriculum used
in the Gemma recipe is backbone-sensitive: it improves Gemma (EM
$30.6\%\!\to\!53.6\%$) but harms Qwen (EM $36.6\%\!\to\!1.2\%$). We therefore train Qwen with shuffled data.

\begin{table}[h]
    \centering
    \caption{Effect of data ordering during training (two epochs).}
    \label{tab:ordering}
    \small
    \begin{tabular}{llccc}
        \toprule
        \textbf{Backbone} & \textbf{Ordering} & \textbf{BLEU} & \textbf{W-ED} & \textbf{EM\,(\%)} \\
        \midrule
        Gemma-270M & curriculum & \textbf{0.978} & \textbf{1.8\%} & \textbf{53.6\%} \\
        Gemma-270M & shuffled   & 0.966 & 10.4\% & 30.6\% \\
        Qwen-0.5B  & curriculum & 0.754 & 28.1\% & 1.2\% \\
        Qwen-0.5B  & shuffled   & \textbf{0.940} & \textbf{7.9\%} & \textbf{36.6\%} \\
        \bottomrule
    \end{tabular}
\end{table}

\section{Adapter Rank and Post-hoc Compression}
\label{sec:eval_rank_compression}
Both experiments in this section are run at the NTP-only stage of Table~\ref{tab:latent_smoothness}, without the latent KL or the codec, with the data ordering fixed to shuffled on both backbones so that the adapter rank is the only variable of the sweep.

\begin{table}[H]
    \centering
    \caption{Teacher-forced cross-entropy loss on the held-out set falls with LoRA rank on both backbones, while Gemma reconstruction peaks at $r{=}8$.}
    \label{tab:rank_ablation}
    \footnotesize
    \setlength{\tabcolsep}{3pt}
    \begin{tabular}{c cccc @{\hskip 6pt} cccc}
        \toprule
        & \multicolumn{4}{c}{\textbf{Gemma-270M}} & \multicolumn{4}{c}{\textbf{Qwen2.5-0.5B}} \\
        \cmidrule(lr){2-5} \cmidrule(lr){6-9}
        \textbf{Rank} & \textbf{Loss}\,($\downarrow$) & \textbf{BLEU} & \textbf{W-ED} & \textbf{EM} & \textbf{Loss}\,($\downarrow$) & \textbf{BLEU} & \textbf{W-ED} & \textbf{EM} \\
        \midrule
        4  & 0.4243 & 0.709 & 37.1\% & 0.4\% & 0.9806 & 0.467 & 94.9\% & 1.4\% \\
        8  & 0.0831 & \textbf{0.966} & \textbf{10.4\%} & \textbf{30.6\%} & 0.0207 & 0.940 & 7.9\% & 36.6\% \\
        16 & 0.0119 & 0.533 & 116\% & 0.4\% & 0.0044 & 0.996 & 0.26\% & 73.8\% \\
        32 & \textbf{0.0060} & 0.832 & 15.8\% & 19.8\% & \textbf{0.0010} & \textbf{0.999} & \textbf{0.07\%} & \textbf{92.8\%} \\
        \bottomrule
    \end{tabular}
\end{table}

\begin{table}[htbp]
    \centering
    \caption{Compressing a trained $K{=}256$ latent to 64 tokens outperforms training $K{=}64$ directly.}
    \label{tab:compression}
    \small
    \begin{tabular}{lcccc}
        \toprule
        \textbf{Model} & \textbf{Latent} & \textbf{BLEU} & \textbf{W-ED} & \textbf{EM} \\
        \midrule
        $K{=}64$ trained directly & 64 & 0.397 & 260\% & 0.0\% \\
        $256{\to}64{\to}256$ post-hoc compressor & 64 & 0.978 & 1.2\% & 34.2\% \\
        $K{=}256$ uncompressed reference & 256 & 0.999 & 0.07\% & 92.8\% \\
        \bottomrule
    \end{tabular}
\end{table}

\paragraph{Teacher-forced loss misranks generators.} Table~\ref{tab:rank_ablation} varies the LoRA rank on both backbones and reports, alongside reconstruction quality, the teacher-forced cross-entropy loss on the held-out set: at every position of the reconstruction segment the model is given the reference prefix and scored on its predicted distribution over the next token. This loss decreases monotonically with rank on \emph{both} backbones, yet Gemma's reconstruction peaks at $r{=}8$ and degrades at $r{=}16$ (BLEU $0.966\!\to\!0.533$, EM $30.6\%\!\to\!0.4\%$). On Qwen the two agree, improving strictly to $r{=}32$. Reconstruction is scored on the model's own greedy output, where every token after the first is conditioned on the tokens the model itself produced, whereas the teacher-forced loss conditions every position on the reference. The two measurements coincide only while the generated prefix matches the reference. A lower loss means the model places more probability on the reference token given the correct context; it does not mean the argmax is correct at every token. Nor does the loss say anything about what happens after a mistake, because a prefix containing the model's own error never occurs during teacher-forced evaluation. The practical consequence is what we act on: teacher-forced cross-entropy on the held-out set is a biased proxy for reconstruction, so every configuration in this paper is selected by generation quality on held-out documents rather than by loss.

\paragraph{Compress a good latent rather than train a small one.} Training the bottleneck directly at $K{=}64$ fails on both backbones (Table~\ref{tab:token_ablation}). On Qwen the reconstructions reach BLEU $0.397$ with a word edit distance of $260\%$ and no document reproduced verbatim: the decoder does not produce a shortened or paraphrased version of the input but drifts into text unrelated to it and several times its length. The question is whether this is a capacity limit of $64$ tokens or a failure to learn such a code through the LLM bottleneck, and Table~\ref{tab:compression} separates the two. We take the trained $K{=}256$ Qwen autoencoder and freeze it entirely. On its latents we fit a Perceiver-style compressor: $64$ learned queries cross-attend to the $256$ latent tokens through six cross-attention blocks (8 heads, each followed by self-attention and a feed-forward layer) to produce a $64{\times}896$ code, and a mirror-image decompressor with $256$ learned queries expands the code back to $256{\times}896$; the two together have $310$M parameters and are trained for one epoch over the cached latents of the $1.4$M training documents. The only objective is the mean squared error between the expanded code and the original latent, computed over all $256{\times}896$ entries after per-coordinate standardization; the compressor never sees a token, the decoder, or a cross-entropy loss. At inference the text is encoded to $256$ tokens, compressed to $64$, expanded back to $256$ and decoded by the untouched decoder. The $64$-token code decodes to BLEU $0.978$, word edit distance $1.2\%$ and $34.2\%$ exact match: not the $92.8\%$ of the uncompressed latent, but a faithful document where direct training at $K{=}64$ produced none. The latent learned at full width is therefore compatible with a global MSE compression to a quarter of its length, whereas the encoder trained at $K{=}64$ never finds a usable code. The difficulty of small $K$ lies in learning the code through the bottleneck, not in the information a $64$-token latent can carry.

\section{Reconstruction by Input Length}
\label{sec:supp_length_buckets}

Table~\ref{tab:token_ablation} reports the latent budget sweep aggregated over the stratified test set. Because the bottleneck holds a fixed $K$ regardless of input length, it is worth asking whether the best $K$ shifts with document length. Table~\ref{tab:length_buckets} breaks the same Gemma models out by input length, using four character-length buckets of the stratified 1024-token test set.

\begin{table}[h]
    \centering
    \caption{Gemma reconstruction (BLEU-4) by input length; buckets are character counts on the stratified test set.}
    \label{tab:length_buckets}
    \small
    \begin{tabular}{ccccc}
        \toprule
        \textbf{Latent tokens} & \textbf{$\leq$1K chars} & \textbf{1-2K} & \textbf{2-3K} & \textbf{3-4K} \\
                               & ($n{=}126$) & ($n{=}129$) & ($n{=}125$) & ($n{=}120$) \\
        \midrule
        64   & 0.573 & 0.298 & 0.293 & 0.342 \\
        128  & 0.336 & 0.162 & 0.184 & 0.184 \\
        \textbf{256}  & \textbf{1.000} & \textbf{0.998} & 0.973 & 0.937 \\
        512  & 0.998 & 0.914 & 0.984 & 0.937 \\
        1024 & 0.677 & 0.995 & \textbf{1.000} & \textbf{0.998} \\
        \bottomrule
    \end{tabular}
\end{table}

The per-bucket optimum does shift with length: $K{=}1024$ is marginally better on the longest inputs but degrades sharply on short ones (BLEU $0.677$ below 1K characters), while $K{=}64$ and $K{=}128$ are capacity-limited throughout. $K{=}256$ is the only setting that remains strong in every bucket, which is why it is optimal in aggregate for a fixed-length autoencoder intended to serve documents of any length. All models here are trained under the same regime, so this analysis varies the evaluation length only; it does not ablate the training length distribution or the maximum input length.

The two degradations have different characters. The $K{=}1024$ loss on short inputs is dominated by a failure to terminate: of the 126 inputs below 1K characters, 37 fall below BLEU $0.5$, and 30 of those run past the end of the document into degenerate filler, with a median reconstruction $15.6\times$ the length of the reference. These cases concentrate on the shortest inputs (mean reference length 210 characters, against 594 for the rest of the bucket), where the latent budget most overshoots the content. $K{=}64$ fails in the opposite way: it terminates correctly and preserves topic and document structure, but substitutes words and phrases throughout, including proper nouns --- the signature of a capacity limit rather than a length mismatch. Table~\ref{tab:length_bucket_examples} gives one example of each.

\begin{table}[h]
    \centering
    \caption{Characteristic failure modes by latent budget and input length. Bold marks corrupted spans.}
    \label{tab:length_bucket_examples}
    \small
    \begin{tabular}{@{}lp{0.85\linewidth}@{}}
        \toprule
        \multicolumn{2}{@{}l}{\textbf{Short input}, 72 characters. BLEU: $K{=}64$ $1.000$, $K{=}256$ $1.000$, $K{=}1024$ $0.489$.} \\
        \midrule
        Reference & We are having a Christmas Party!! Come join us with some Holiday Cheer!! \\
        $K{=}256$ & We are having a Christmas Party!! Come join us with some Holiday Cheer!! \\
        $K{=}1024$ & We are having a Christmas\textbf{, however} Come join us with some Holiday\textbf{,}!! \\
        \midrule
        \multicolumn{2}{@{}l}{\textbf{Long input}, 3{,}269 characters (excerpt). BLEU: $K{=}64$ $0.265$, $K{=}256$ $0.985$, $K{=}1024$ $1.000$.} \\
        \midrule
        Reference & \ldots said he has no regrets about taking a break from his successful academic career for the rough and tumble of a Greek election campaign, his second in a year. Eleftheriadis is running for To Potami\ldots \\
        $K{=}256$ & \ldots said he has no regrets about taking a break from his successful academic career for the rough and tumble of a Greek election campaign, his second in a year. Eleftheriadis is running for To Potami\ldots \\
        $K{=}64$ & \ldots said he has no regrets about taking \textbf{what} break from his successful academic career for the rough and tumble of a Greek election campaign, his second in a year. \textbf{Elefther is the a running change} To Potami\ldots \\
        \bottomrule
    \end{tabular}
\end{table}

\section{ICAE-style Autoencoder with Gemma-270M}
\label{sec:supp_icae}

\begin{table}[t]
    \centering
    \caption{Reconstruction quality for ICAE-style training on Gemma-270M}
    \label{tab:gemma_icae}
    \begin{tabular}{lcccc}
        \toprule
        \textbf{\# Tokens} & \textbf{Curriculum Learning} &  \textbf{BLEU} ($\uparrow$) & \textbf{PPL} ($\downarrow$) & \textbf{BERT} ($\uparrow$) \\ 
        \midrule
        256 & No & 0.014 & 10.4 & 0.785 \\
        256 & Yes & 0.087 & 267410 & 0.801 \\
        512 & Yes & 0.393 & 204.1 & 0.922 \\
        \bottomrule
    \end{tabular}
\end{table}

To evaluate the utility of the LLMAE architecture, we implemented a baseline following the In-context Autoencoder (ICAE) \cite{ge2023context} framework. This baseline utilizes a soft prompting paradigm where the K latent tokens serve as a compressed ``memory tokens'' prefix to the input tokens for autoregressive decoding. We instantiated this model using the same Gemma-270M backbone and training data we use for LLMAE. 

Our experiments indicate that this soft prompting approach is insufficient for high-fidelity text autoencoding for the lightweight Gemma backbone. This performance bottleneck remained non-trivial with or without employing curriculum learning (sorting training samples by sequence length). The inability of the ICAE-style prefix to recover fine-grained semantic and syntactic details suggests that simply treating latents as a soft prompt does not provide the necessary structural guidance for long-form reconstruction, at least in the Gemma 270M backbone. We provide the reconstruction evaluations in Table~\ref{tab:gemma_icae}.

\section{Exact Reconstruction Fidelity Metrics}
\label{sec:supp_fidelity}
\label{sec:supp_exact_fidelity}

\begin{table*}[t]
\centering
\caption{Perplexity (GPT-2-Large, $\downarrow$) and BERTScore F1 ($\uparrow$) for the settings of Table~\ref{tab:reconstruction}. Reported in the appendix: BERTScore truncates long inputs, and degenerate reconstructions can attain lower perplexity than the source text.}
\label{tab:reconstruction_pplbert}
\small
\setlength{\tabcolsep}{4pt}
\begin{tabular}{@{} l cc @{\hskip 6pt} cc @{\hskip 6pt} cc @{}}
\toprule
& \multicolumn{2}{c}{\textbf{C4-News-Strat.}} & \multicolumn{2}{c}{\textbf{OpenWebText}} & \multicolumn{2}{c}{\textbf{CreationMMBench}} \\
\cmidrule(lr){2-3} \cmidrule(lr){4-5} \cmidrule(lr){6-7}
\textbf{Method} & \textbf{PPL} & \textbf{BERT} & \textbf{PPL} & \textbf{BERT} & \textbf{PPL} & \textbf{BERT} \\
\midrule
Source Text & 30.8 & -- & 27.5 & -- & 16.9 & -- \\
ICAE & 36.8 & 0.981 & 30.0 & 0.963 & 17.7 & 0.983 \\
COSMOS & 66.9 & 0.979 & 32.6 & 0.982 & 23.3 & 0.951 \\
LLMAE-Gemma & 30.9 & 0.999 & 27.6 & 0.999 & 17.1 & 0.998 \\
LLMAE-Qwen & 31.2 & 1.000 & 28.8 & 0.999 & 17.4 & 0.999 \\
\midrule
Source Text & 24.1 & -- & 21.9 & -- & 13.3 & -- \\
ICAE & 30.2 & 0.968 & 23.0 & 0.948 & 14.4 & 0.970 \\
LLMAE-Gemma & 24.4 & 0.999 & 22.1 & 0.998 & 13.4 & 0.998 \\
\quad w/o Codec & 26.9 & 0.993 & 24.7 & 0.993 & 14.6 & 0.994 \\
LLMAE-Qwen & 24.2 & 1.000 & 22.0 & 1.000 & 13.6 & 1.000 \\
\quad w/o Codec & 24.2 & 1.000 & 22.0 & 1.000 & 13.4 & 1.000 \\
\bottomrule
\end{tabular}
\end{table*}

Table~\ref{tab:reconstruction_pplbert} reports perplexity (GPT-2-Large) and BERTScore F1 for the settings of Table~\ref{tab:reconstruction}. To complement the BLEU-4, exact-match, and edit-distance results there, we report direct, token-level fidelity metrics computed on the same model outputs and test sets. All metrics operate on whitespace-delimited words, making them independent of any model-specific subword tokenizer and therefore directly comparable across the Gemma- and Qwen-based LLMAE instantiations, the Mistral-based ICAE, and the BERT-based COSMOS. We report the per-category error analysis in Table~\ref{tab:category_breakdown}.

\paragraph{Metric definitions.}
For each test sample, let $\mathbf{w} = (w_1, \dots, w_n)$ denote the reference word sequence and $\hat{\mathbf{w}} = (\hat{w}_1, \dots, \hat{w}_m)$ the reconstruction word sequence, obtained by whitespace splitting.
\begin{itemize}
  \item \textbf{Exact match (EM):} the fraction of samples for which the reconstruction equals the reference in full after whitespace normalization (runs of whitespace collapsed to single spaces). This is a strict, binary, whole-document criterion with no partial or prefix credit: a sample scores $1$ only if the reconstruction is identical to the reference from the first character to the last, and a single incorrect character anywhere in the document yields $0$. EM is therefore the fraction of documents reproduced verbatim end-to-end; an error in the final word is penalized exactly as heavily as an error in the first.
  \item \textbf{Word edit distance (W-ED):} the Levenshtein distance between $\mathbf{w}$ and $\hat{\mathbf{w}}$ (minimum number of word insertions, deletions, and substitutions), normalized by $n$ and averaged over samples.
\end{itemize}

\paragraph{Category breakdown.}
For the per-category analysis in Table~\ref{tab:category_breakdown}, every \emph{reference} word is assigned to exactly one category, in priority order: \emph{number} (contains a digit), \emph{punctuation} (no alphanumeric characters), \emph{capitalized} (begins with an uppercase letter; a proxy for named entities that conservatively also includes sentence-initial words), \emph{rare} (lowercase word occurring at most twice among the references of that evaluation set), and \emph{common} (all remaining words). A reference word counts as an error if the alignment marks it as substituted or deleted. The per-category error rate is the number of errored words divided by the total reference words in that category, pooled over all samples.

\begin{table}[t]
    \centering
    \caption{Word-error rate ($\downarrow$) of LLMAE by reference-word category on the C4-News-Stratified evaluation sets. Errors show no systematic concentration in numbers, capitalized words (named-entity proxy), punctuation, or rare tokens. On the same categories, the baseline error rates range from 0.12-0.36 (ICAE) and 0.02-0.38 (COSMOS).}
    \label{tab:category_breakdown}
    \small
    \begin{tabular}{lccccc}
        \toprule
        \textbf{Setting} & \textbf{Number} & \textbf{Punctuation} & \textbf{Capitalized} & \textbf{Rare} & \textbf{Common} \\
        \midrule
        C4-News-Stratified (512)  & 0.0011 & 0.0028 & 0.0019 & 0.0009 & 0.0004 \\
        C4-News-Stratified (1024) & 0.0069 & 0.0028 & 0.0046 & 0.0057 & 0.0029 \\
        \bottomrule
    \end{tabular}
\end{table}

\paragraph{Interpretation.}
Exact match compounds sharply with document length even at fixed per-token fidelity: a single imperfect word among $\sim$500 zeroes the match, however low the word edit distance (Table~\ref{tab:reconstruction}). The per-document error distribution of the Gemma LLMAE confirms that misses are near-misses, consistently across all three domains: at 1024 tokens, 67-71\% of documents reconstruct verbatim, 84-87\% are within 2 word edits of the reference, 94-95\% are within 5, and the median non-exact document differs in only 2-3 words. The baselines' low EM arises from qualitatively different error modes: ICAE's errors are content substitutions whose rate grows with position in the document (on every domain the word-error rate in the second half exceeds the first half, e.g.\ $14.7\%$ vs.\ $22.8\%$ on C4-News-Stratified at 1024 tokens), consistent with the length degradation in Table~\ref{tab:reconstruction}, whereas COSMOS almost never reproduces a document verbatim (94-100\% of documents contain at least one word error) largely because its BERT-tokenizer decoding pipeline inserts spaces around intra-word punctuation and drops line breaks when detokenizing (e.g., \emph{op-ed}$\rightarrow$\emph{op - ed}, \emph{2014's}$\rightarrow$\emph{2014 ' s}, \emph{27.8}$\rightarrow$\emph{27.\ 8}). The characters are largely preserved but word boundaries are not, so word-level metrics register errors on nearly every document.

\section{Method Details}
\label{sec:supp_architecture}
\subsection{LLMAE}
We describe the Gemma recipe; the Qwen2.5-0.5B LLMAE follows the same recipe and differs only in the entries for which Table~\ref{tab:llmae_hyperparameters} lists two values (latent layer 15, LoRA rank 32 with alpha 64, shuffled data ordering, and 23.8M trainable parameters). LLMAE trains Gemma-3-270M-PT (not instruction-tuned) with LoRA on all attention projections for 2 epochs over 1.4M stratified text samples (1M Pile-uncopyrighted + 400K C4-RealNewsLike). The reconstruction test sets contain 500 held-out samples per dataset and length setting, with the exception of CreationMMBench at 512 tokens (311 samples): the source benchmark provides 765 GPT-4o reference answers in total, and we retain only those within the 512-token limit under all evaluated tokenizers (Gemma, BERT, and LLaMA) so that every baseline is evaluated on identical texts. The C4-News-Stratified test sets are constructed to be explicitly disjoint from the training corpus: every candidate text is checked by content hash against all 1.4M training samples and excluded on any match. The other two sources cannot overlap with the training data by construction: OpenWebText uses the held-out test split shared by COSMOS, and the CreationMMBench references are GPT-4o-generated text from an external benchmark. Within each set, the evaluated subset is a fixed deterministic draw (a seeded permutation of the set, truncated to the reported sample count), so every method in Table~\ref{tab:reconstruction} is scored on identical texts. Across the six settings the evaluated documents span up to 962 tokens with a mean of 438; the 1024-token ceiling is set by the least token-efficient tokenizer among those evaluated rather than by LLMAE. On Gemma, LLMAE utilizes LoRA adapters (737K parameters), 256 latent token embeddings (164K parameters), and an additive Codec (9.8M), totalling 10.7M trainable parameters. The input layout is \texttt{[input tokens $|$ 256 latent tokens $|$ input copy]} with the LM loss computed only on the reconstruction segment. Texts are sorted by length (curriculum learning), output logits are regularized with KL toward the frozen reference model, and Gaussian noise ($\sigma{=}0.1$) is added to the latent activations at layer 11 during training to encourage a smooth latent space. The model is then fine-tuned for 2 additional epochs either with the latent KL toward $\mathcal{N}(0,I)$ on the raw latent (the + KL row of Table~\ref{tab:latent_smoothness}) or, for our final model, with a lightweight additive Codec inserted at layer 11 whose refined latents are regularized toward $\mathcal{N}(0,I)$; the coefficients of each stage are listed in Table~\ref{tab:llmae_hyperparameters}. The full parameter configuration is listed in Table~\ref{tab:llmae_hyperparameters}.
\begin{table}[h]
\centering
\caption{LLMAE hyperparameters for both backbones. Values shared by both are given once.}
\label{tab:llmae_hyperparameters}
\footnotesize
\setlength{\tabcolsep}{4pt}
\begin{tabular}{lll}
\toprule
\textbf{Parameter} & \textbf{Gemma-3-270M} & \textbf{Qwen2.5-0.5B} \\
\midrule
Input text length & \multicolumn{2}{l}{1024 tokens} \\
Latent tokens & \multicolumn{2}{l}{256} \\
Latent layer & 11 (12 of 18) & 15 (16 of 24) \\
Latent dimension & 640 & 896 \\
Trainable parameters & 10.7M & 23.8M \\
\quad LoRA / latent embeddings / codec & 0.74M / 0.16M / 9.8M & 4.3M / 0.23M / 19.3M \\
\midrule
\multicolumn{3}{l}{\textit{LoRA}} \\
Rank $r$ / alpha & 8 / 16 & 32 / 64 \\
Dropout & \multicolumn{2}{l}{0.05} \\
Target modules & \multicolumn{2}{l}{\texttt{q\_proj, k\_proj, v\_proj, o\_proj}} \\
Saved modules & \multicolumn{2}{l}{\texttt{embed\_tokens}} \\
\midrule
\multicolumn{3}{l}{\textit{Training}} \\
Dataset & \multicolumn{2}{l}{1M Pile-uncopyrighted + 400K C4-RealNewsLike} \\
Total samples & \multicolumn{2}{l}{1{,}400{,}000} \\
Epochs & \multicolumn{2}{l}{2} \\
Batch size & \multicolumn{2}{l}{32 (4 per GPU $\times$ 8 GPUs)} \\
Gradient accumulation steps & \multicolumn{2}{l}{4 (effective batch 128)} \\
Learning rate & \multicolumn{2}{l}{$5 \times 10^{-5}$} \\
Latent noise std ($\sigma$) & \multicolumn{2}{l}{0.1} \\
Data ordering & curriculum (sorted by length) & shuffled \\
Precision & \multicolumn{2}{l}{bfloat16} \\
\midrule
\multicolumn{3}{l}{\textit{Regularization}} \\
Reference-model KL coefficient & \multicolumn{2}{l}{$1 \times 10^{-4}$} \\
Latent KL coefficient (+KL stage, no codec) & \multicolumn{2}{l}{$1 \times 10^{-3}$} \\
Latent KL coefficient (with codec) & \multicolumn{2}{l}{0; the codec KL below regularizes the codec embedding} \\
\midrule
\multicolumn{3}{l}{\textit{Codec fine-tuning}} \\
Additional epochs & \multicolumn{2}{l}{2} \\
Codec KL weight & \multicolumn{2}{l}{$1 \times 10^{-3}$} \\
Codec learning rate & \multicolumn{2}{l}{$1 \times 10^{-4}$} \\
Data ordering & \multicolumn{2}{l}{shuffled} \\
\bottomrule
\end{tabular}
\end{table}
\subsection{Latent Diffusion for Image Captioning}
The score network and diffusion process are based on the architecture utilized by COSMOS \cite{meshchaninov2025cosmos}, incorporating three primary modifications to align with the LLMAE latent space. First, the architecture is resized to the LLMAE latent dimensionality: $256 \times 640$ for the Gemma LLMAE (10 attention heads of size 64) and $256 \times 896$ for the Qwen2.5 LLMAE (14 heads of size 64), with a SwiGLU FFN width of 2560 in both cases. Second, we introduce image conditioning by projecting SigLIP features through a shared linear layer ($1152 \to 640$ or $1152 \to 896$). These features are integrated via a per-block cross-attention sublayer utilizing RMSNorm pre-norms on both query and key/value sequences. A learnable scalar residual gate, initialized to zero, is applied to these sublayers to allow the model to begin as a purely unconditional denoiser and gradually learn to incorporate image signals. Third, classifier-free guidance (CFG) is implemented by replacing the image conditioning with a learned null embedding with a $0.1$ probability during training, enabling CFG extrapolation during inference. All other structural and procedural components—including the Tanh VP-SDE schedule ($d=5$), U-Net skip connections, $x_0$-prediction parameterization, self-conditioning, per-dimension latent normalization, and the Euler reverse solver—remain identical to the original COSMOS diffusion implementation. In our primary configuration, the autoencoder is our pre-trained LLMAE (278M; Gemma-3-270M backbone with a 9.8M Codec) operating on a compact 256×640 latent space, and the score network is a 12-layer transformer with hidden dimension 640 (111M parameters), giving a total inference footprint of ${\sim}819$M parameters with only the score network trained. For the Qwen2.5 LLMAE (494M; Qwen2.5-0.5B backbone with a 19.3M Codec) the score network widens to hidden dimension 896 (183M parameters), for a total of ${\sim}1{,}105$M with 183M (16.6\%) trained; all other settings are identical and the two models are trained and evaluated under the same protocol. Replacing LLMAE with the COSMOS autoencoder (356M; 208M encoder, 149M decoder) expands the latent space to 512×768, which requires a wider score network (hidden dimension 768, 159M parameters) to match; the total system grows to ${\sim}943$M parameters, with the trained fraction increasing from 111M to 159M (13.6\% to 16.9\% of total). The full parameter configuration is listed in Table~\ref{tab:captioning_hyperparams}.

\begin{table}[h]
\centering
\caption{Hyperparameters for latent diffusion captioning, per LLMAE backbone. Values shared by both are given once.}
\label{tab:captioning_hyperparams}
\footnotesize
\setlength{\tabcolsep}{4pt}
\begin{tabular}{lll}
\toprule
\textbf{Parameter} & \textbf{Gemma LLMAE} & \textbf{Qwen2.5 LLMAE} \\
\midrule
\multicolumn{3}{l}{\textit{Score network}} \\
Latent space (tokens $\times$ dim) & $256 \times 640$ & $256 \times 896$ \\
Hidden size $\times$ layers $\times$ heads & $640 \times 12 \times 10$ & $896 \times 12 \times 14$ \\
FFN intermediate size & \multicolumn{2}{l}{2560} \\
Trainable parameters & 111M & 183M \\
Total inference footprint & 819M & 1{,}105M \\
\midrule
\multicolumn{3}{l}{\textit{Image conditioning}} \\
Vision encoder & \multicolumn{2}{l}{SigLIP SO400M patch14-384 (frozen, 428M params)} \\
Encoder output & \multicolumn{2}{l}{729 patches $\times$ 1152 dims} \\
Conditioning projection & Linear($1152 \to 640$) & Linear($1152 \to 896$) \\
Cross-attention gate init & \multicolumn{2}{l}{0.0} \\
CFG null embedding & \multicolumn{2}{l}{Learnable, zero-init} \\
Cond drop probability & \multicolumn{2}{l}{0.1} \\
CFG scale (inference) & \multicolumn{2}{l}{3.0} \\
\midrule
\multicolumn{3}{l}{\textit{Diffusion process}} \\
Noise schedule & \multicolumn{2}{l}{Tanh VP-SDE, $d = 5$} \\
$x_0$-prediction & \multicolumn{2}{l}{Yes} \\
Self-conditioning & \multicolumn{2}{l}{Yes} \\
Latent normalization & \multicolumn{2}{l}{Per-(token, dim) mean/std} \\
Inference steps & \multicolumn{2}{l}{250} \\
\midrule
\multicolumn{3}{l}{\textit{Training}} \\
Batch size & \multicolumn{2}{l}{64 per GPU $\times$ 8 GPUs} \\
Training iterations & \multicolumn{2}{l}{226,000} \\
Learning rate & \multicolumn{2}{l}{$2 \times 10^{-4}$ (2M-step cosine to $2 \times 10^{-5}$, halted at 226K; constant in effect)} \\
Warmup steps & \multicolumn{2}{l}{2,000} \\
Optimizer & \multicolumn{2}{l}{AdamW ($\beta_1=0.9, \beta_2=0.98$, weight decay $10^{-5}$)} \\
EMA decay (tracked) & 0.9999 & 0.99995 \\
Inference weights & \multicolumn{2}{l}{Raw score-network weights (EMA not applied)} \\
Precision & \multicolumn{2}{l}{float32} \\
\bottomrule
\end{tabular}
\end{table}

\end{document}